\documentclass[11pt]{article}

\usepackage[margin=1in]{geometry}
\usepackage{amsmath,amssymb,amsfonts,amsthm}
\usepackage{graphicx}
\usepackage{hyperref}
\usepackage{acronym}

\title{Multifractal Recalibration of Neural Networks for Medical Imaging Segmentation}

\author{%
  Miguel L.~Martins\\
  Department of Computer Science, University of Porto \& INESC-TEC\\
  \texttt{miguel.l.martins@inesctec.pt}
  \and
  Miguel T.~Coimbra\\
  Department of Computer Science, University of Porto \& INESC-TEC\\
  \texttt{mcoimbra@fc.up.pt}
  \and
  Francesco Renna\\
  Department of Computer Science, University of Porto \& INESC-TEC\\
  \texttt{francesco.renna@fc.up.pt}
}

\usepackage{acronym}
\acrodef{CNN}{convolutional neural network}
\acrodef{ANN}{artificial neural network}
\acrodef{DNN}{deep neural network}
\acrodef{ReLU}{rectified linear unit}
\acrodef{MLP}{multilayer perceptron}
\acrodef{DNN}{deep neural network}
\acrodef{FCN}{fully convolutional network}
\acrodef{SRM}{style based recalibration}
\acrodef{SVM}{support vector machine}
\acrodef{FCA}{frequency channel attention}
\acrodef{SE}{squeeze-and-excitation}
\acrodef{SSR}{singularity strength recalibration}
\acrodef{MFS}{{multifractal spectrum}}
\acrodef{SE}{squeeze-and-excitation}
\acrodef{OLS}{ordinary least-squares}
\acrodef{WTMM}{{Wavelet transform modulus-maxima}}
\acrodef{MFDFA}{{Multifractal detrended fluctuation analysis}}
\acrodef{GAP}{global average pooling}
\acrodef{MRI}{magnetic resonance imaging}
\acrodef{DBC}{{differential box counting}}
\acrodef{DFA}{{detrended fluctuation analysis}}
\acrodef{FLOPS}{floating point operations per second}
\acrodef{PCA}{principle component analysis}
\acrodef{NAS}{neural architectural search}
\acrodef{EPT}{excitation PCA threshold}

\newenvironment{keywords}{%
  \begingroup\small\noindent\textbf{Keywords.} }{\par\endgroup}
\newenvironment{AMS}{%
  \begingroup\small\noindent\textbf{AMS subject classifications.} }{\par\endgroup}

\theoremstyle{plain}
\newtheorem{theorem}{Theorem}[section]

\theoremstyle{definition}
\newtheorem{definition}[theorem]{Definition}

\theoremstyle{remark}
\newtheorem{remark}[theorem]{Remark}

\newcommand{\qedwhite}{\hfill\ensuremath{\Box}}

\date{} 

\begin{document}

\maketitle

\begin{abstract}
    Multifractal analysis has been pivotal to uncover the regularities of several self-seeding phenomena in various scientific measurements. However, although methods characterizing the so-called \emph{multifractal spectrum} (MFS) received a considerable amount of attention in classical computer vision, its role in the era of modern deep learning has been relatively limited. In fact, some end-to-end approaches emerged, mainly for the task of texture recognition. These new methods rely on computationally heavy pooling operations, or aggressive decimation of the feature space in order to enable training and inference to be feasible in time, which are unsuitable for tasks such as semantic segmentation.
    
    With this motivation, we set forth two new inductive priors that address these limitations -- Monofractal and Multifractal Recalibration. We leverage the relationships between the probability mass of the exponents and their MFS to build a statistical description of each embedding of the encoder of the network. We thus formulate our solutions as channel-attention functions in the context of convolutional neural networks.  
    
    We build an experimental framework centered around the U-Net and show that Multifractal recalibration can lead to substantial improvements over a baseline augmented with other well-established channel attention functions that also describe each channel in terms of higher-order statistics. Due to the proven effectiveness of multifractal analysis in capturing pathological regularities, we conduct our experiments over three public datasets from diverse medical modalities: ISIC18 (dermoscopy), Kvasir-SEG (endoscopy), and BUSI (ultrasound).  

An empirical analysis also reveals new insights into the dynamics of these attention layers. We find that excitation response does not get increasingly specialized with encoder depth in the U-Net due to its skip connections, and that its effectiveness may be linked to global statistics of their instance-variability.  
\end{abstract}



\section{Introduction}
Pathological phenomena  may be described as a \textit{state} of a physiological system that evolves in a self-seeding, cascading fashion \cite{karperien2016multifractal}. These states are observed as complex patterns in medical imaging modalities, whose regularities may be disentangled through the lens of \textit{Fractal Geometry}, the field of mathematics centered around the study of mathematical structures that have fine-structure, i.e., detail at all scales \cite{falconer2004fractal}. This degree of scale-invariance may be captured in terms of a single (monofractal) or multiple (multifractal) scaling exponents that embody the \textit{dimension} of the structure~\cite{evertsz1992multifractal}. This \textit{self-similarity} in distribution across scale is especially useful to describe textures as stationary stochastic processes~\cite{chaudhuri1995texture, xia2006morphology, xu2021encoding, chen2021deep}.

In Medical Imaging, both structural and surface-level regularities may be leveraged \cite{karperien2016multifractal}.  
\begin{figure}
    \centering
    \includegraphics[width=1.00\columnwidth]{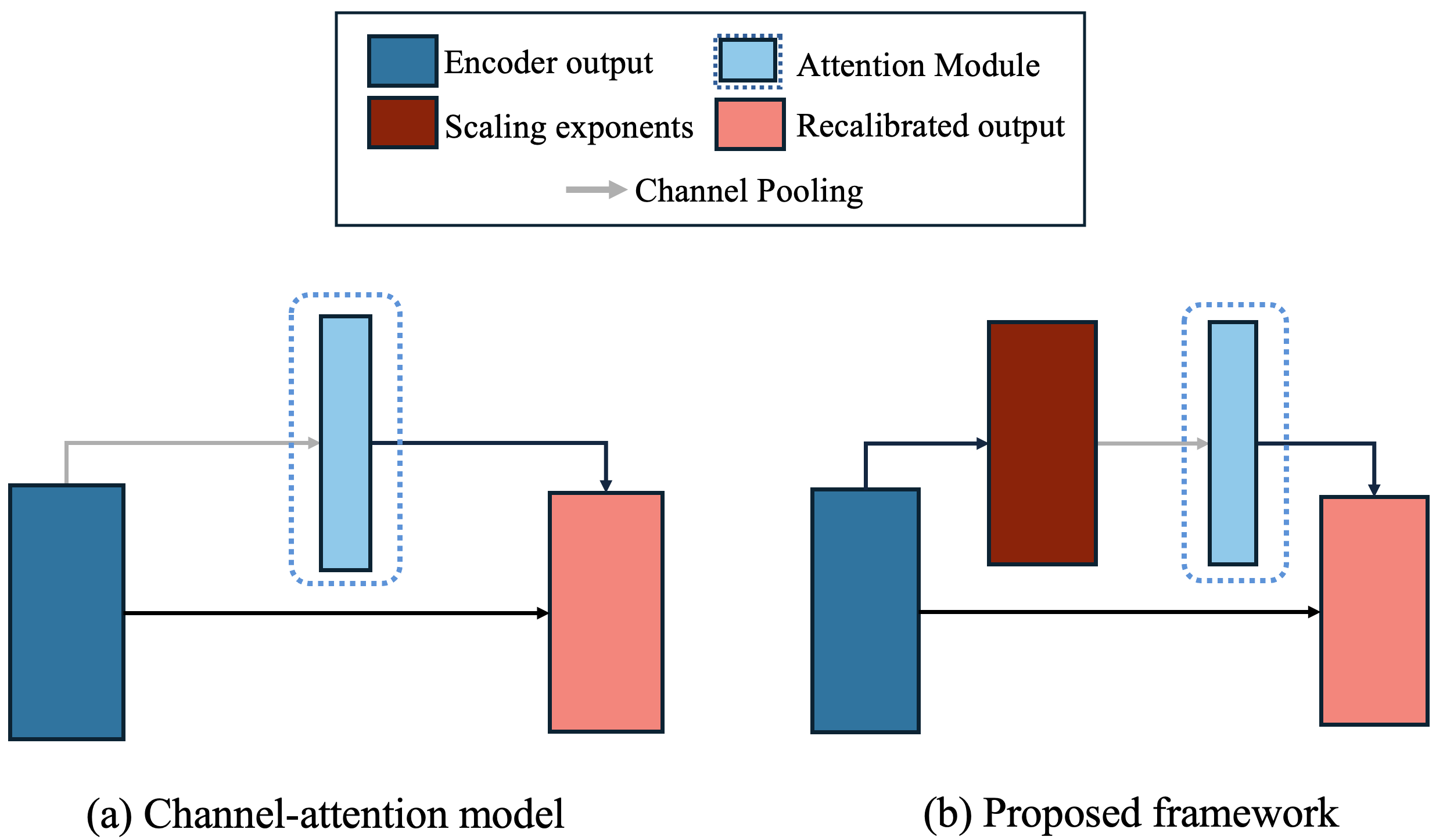}
    \caption{(a) The typical channel attention model where some statistics is pooled directly from the encoder output. (b) Our proposed approaches use statistics derived from the scaling exponents of the features maps. }
    \label{fig:proposed_framework}
\end{figure}
Examples range from histology \cite{atupelage2012multifractal, nawn2020multifractal, roberto2021fractal}, endoscopy \cite{hafner2015local, Maria2023}, dermoscopy~\cite{piantanelli2005fractal, chatterjee2018optimal}, to \ac{MRI}~\cite{korchiyne2012medical, islam2013multifractal, lahmiri2017glioma} and many others (see~\cite{lopes2009fractal,karperien2016multifractal} and references therein). On the other hand, Fractal Geometry has received limited attention in the deep learning era, especially in the design of end-to-end architectures. This trend recently changed, and  some successful approaches have been proposed for texture classification~\cite{xu2021encoding, chen2021deep, chen2024mfen} and few-shot learning~\cite{zhou2023fractal}. Bridging this gap for semantic segmentation is the main contribution of this paper, under the assumption that the inductive prior of a \ac{CNN} encoder carries information about the image's scaling laws across its layers.  

Grounded in multifractal analysis, we propose two new attention functions designed around the U-Net \cite{ronneberger2015u} , the main specialist model alternative to current foundational approaches \cite{isensee2021nnu, isensee2024nnu, ma2024segment}. These functions describe cross-channel dynamics by deriving single or multiple statistics over the local scaling exponents of each filter (see Figure~\ref{fig:proposed_framework}). We show substantial improvements over several attention functions~\cite{roy2018recalibrating, lee2019srm, qin2021fcanet} in terms of segmentation performance in three public medical image segmentation datasets.
During this investigation we also conduct an empirical analysis in an effort to shed some light on the reason behind the disparity of performance between different statistical channel functions. Our findings suggest that intuitions held by the community behind the mechanisms that govern these layers\cite{hu2018squeeze, roy2018recalibrating, pereira2019adaptive, guo2022attention} do not seem to hold in the case of semantic segmentation. Briefly, our experiments suggest that these functions do not seem to learn to ``filter non-informative channels''. Inspection of their response dynamics are indeed antithetical to this assumption: a certain range of instance variability is actually desirable. 

\subsection{Contributions}
Current end-to-end architectures  \cite{xu2021encoding, chen2021deep, chen2024mfen} that leverage multifractal primitives (described in Section \ref{sec:rel_work:modern}) are engineered for recognition, and most draw parallels with the underlying formalism in a very ad hoc fashion. We only identified the fractal encoder in \cite{xu2021encoding} to be close to a true multifractal prior. Moreover, these methods rely on computationally expensive fusion approaches that require spatial or channel upsampling and/or downsampling  \cite{xu2021encoding, zhou2023fractal, chen2024mfen}, aggressive shape normalization across spatial and channel dimensions~\cite{chen2021deep}, and inefficient~\cite{chen2024mfen} or repeated computations~\cite{chen2021deep} of extensions of the \ac{DBC} algorithm. 

We address these issues by deriving statistics over the local scaling exponents with a theoretically guided rationale. This work has a special focus on the U-Net, but note that the methods set forth may be readily applied to any \ac{DNN} architecture.  
We then draw inspiration from statistical channel attention functions \cite{hu2018squeeze, roy2018recalibrating, lee2019srm, qin2021fcanet, guo2022attention}, but we recalibrate the output of the network as a function of the scaling exponents, instead of using the direct responses of each encoder block.

This paper also extends very preliminary empirical work where we presented an early version of Monofractal recalibration \cite{martins2024singularity}. However, the experimental setup has been substantially upgraded with more extensive and rigorous statistical analyses, more datasets, better training and cross-validation routines, more baseline architectures \cite{roy2018recalibrating, lee2019srm, qin2021fcanet} and ablation experiments, and a more detailed analysis of how these attention mechanisms behave. Finally, we propose an entire new attention module that displays superior performance in general: \emph{Multifractal recalibration}. All of the introduced inductive priors are also studied from a principled perspective through the lens of multifractal anaylsis.

Our main contributions are summarized as follows\footnote{Our code is available at \url{https://github.com/miguelmartins/multifractal-recalibration/}.}:

\begin{enumerate}
    \item An efficient end-to-end way to compute the local scaling exponents adequate for dense prediction tasks.
    \item Two novel attention functions thoeretically grounded in multifractal analysis.
    \item An experimental analysis of other statistically informed recalibration strategies beyond \ac{SE} \cite{roy2018recalibrating}, such as \ac{SRM} \cite{lee2019srm} and \ac{FCA}~\cite{qin2021fcanet} for the purposes of medical image segmentation.
\end{enumerate}

\subsection{Paper structure}
\label{sec:struacture}
The remainder of the paper is structured as follows. Section \ref{sec:background} provides a detailed description of the application of Fractal Geometry in Computer Vision, Medical Imaging, and with regards to its latest advancements in the deep learning era. 
In Section \ref{sec:methods:perliminaries} we conduct a background review of the core concepts and set forth the formalism that will contribute to our proposed approaches, presented in Section \ref{sec:methods}. The performance of Monofractal and Multifractal recalibration is measured on three public datasets in Section \ref{sec:experiments}, alongside an empirical anylsis of the excitation responses. Finally, in Section \ref{sec:experiments:discussion}, we identify limitations and future work, after highlighting our main theoretical and experimental findings. 
\section{Related Work}
\label{sec:background}
This section provides a review of the prior art, with initial context from classical computer vision, and then re-contextualized within more contemporary natural and medical imaging. Finally, we study how Fractal Geometry has been leveraged to develop novel inductive biases for deep learning methods. 

\subsection{Fractals in Classical Computer Vision}
\label{sec:rel_work:classical_cv}
The first applications of Fractal Geometry in Computer Vision date to back the 1980s \cite{barnsley1988science}. In this era, Pentland seminally argued that since the intensity of an image is a function of the angle between the illuminant and the surface normal, then if the normals follow a Fractal Brownian distribution and the perspective is kept constant, then the fractal dimension is carried to a digital photograph \cite{pentland1983fractal}. The introduction of the \ac{DBC} algorithm \cite{sarkar1994efficient} was pivotal to increase adoption of fractal geometry in the Computer Vision community, since it allowed for efficient estimation of the number of boxes covering a 3D surface (e.g., a gray-scale image).

The fractal dimension has been used extensively as a descriptor for texture \cite{chaudhuri1995texture, xu2009viewpoint, varma2007locally}. Quantities related to the more general \emph{set of generalized dimensions} \cite{xia2006morphology} embedded within morphological models and clustering approaches were also demonstrated to be effective in the semantic segmentation of textural attributes, such as mosaics of textures from the Brodatz album \cite{brodatz1966textures}.  Other classical approaches for multifractal segmentation can be found in the references of \cite{xia2006morphology, varma2007locally}.

 In \cite{vehel1994multifractal}, Vehel et al. propose  multifractal formalism as an alternative to the traditional theory of filtering. They noted that the differentiation operation commutes with convolution, and so edges and regions of interest were found after the image was smoothed through convolution with some function $g$. Thus, by leveraging Multifractal formalism, one could potentitally assign each point to a specific primitive (e.g., an edge) depending on the scaling exponent around its (local) neighbourhood, and each resulting on a \emph{level set partition} that could be associated with the output of the convolution of the image with said filter $g$. 

Vehel et al. also introduced the important idea of capturing this scaling behaviour as a function of several different \emph{capacity functions} (such as the maximum, minimum, sum, etc.) in order to mitigate the noise sensitivity these methods display in practice. This inspired several contributions, where quantities related either to a set of fractal dimensions \cite{varma2007locally}, or even the so called \ac{MFS} \cite{evertsz1992multifractal, xu2009viewpoint}, $f(\alpha)$, were computed for a set of general filters, each defining a \emph{measure} of the original (gray-scale) image.

Most of these early approaches were tailored towards gray-scale images. Inspired by multifractal theory \cite{barnsley1988science} Ivanovici et al. proposed to measure the Fractal Geometry of RGB images \cite{ivanovici2010fractal, ivanovici2013color}, by extending the concepts of Box dimension to 5-dimensional metric spaces. 

\subsection{Fractals in Medical Imaging}
\label{sec:rel_work:medical}
Physiologic systems have been described as inherently (multi)fractal since they evolve in a self-seeding, cascading fashion \cite{karperien2016multifractal}. In fact, the theory of fractal and self-organizing structures has been proposed as a fundamental design principle of life itself \cite{kurakin2011self}. \textit{Fractalomics}~\cite{losa2009fractal} has been proposed to leverage statistical quantities based in fractal geometry due to its effectiveness at describing quantitatively morphological cellular structure complexity of cellular and biological tissues. Pathologies such as tumors have also exhibited fractal properties in both space and time~\cite{molski2008tumor}. For example, in dermoscopy, the fractal dimension was used to associate the regularity of the boundary of dermatological lesions with their severity~\cite{piantanelli2005fractal}. This was eventually extended to include lesion texture to enhance classification~\cite{chatterjee2018optimal}. It has been used to segment calcifications or regress bounding boxes in lesions captured in breast ultrasounds~\cite{stojic2006adaptation, yap2008novel}. It was also adopted for~\ac{MRI} to segment or classify tumors~\cite{korchiyne2012medical, islam2013multifractal, lahmiri2017glioma}. Its power to derive texture features significantly contributed to the increase of classifier accuracy in several histopathology modalities~\cite{atupelage2012multifractal, nawn2020multifractal, roberto2021fractal} as well as polyp and mucosal characterization in endoscopy~\cite{hafner2015local, Maria2023}. 

In fact, fractal and multifractal methods intersect with virtually all medical modalities in such a profound way that a thorough review of the art would require its own publication, so we refer the interest reader to \cite{lopes2009fractal, west2012fractal}. In the case of medical images, although the number of use-cases is broad, the type of techniques is relatively narrow, and do not differ significantly from the rest of the computer vision literature.
\subsection{Fractals in Modern Computer Vision}
\label{sec:rel_work:modern}
Fractal geometry has received minimal attention as a design component for end-to-end deep learning architectures until relatively recently. In fact, prior to this paper, we found no end-to-end architectures for semantic segmentation. One of the main limiting factors is the expensive (and often non-differentiable, in the case of DBC) multi-scale analysis required to compute the local scaling exponents. Moreover, it is not entirely clear how fractal geometry fits in architectures such as \ac{CNN}s, since they can learn arbitrary functions, and are thus not constrained to behave strictly as a set of Wavelet basis functions, as typically required in multifractal formalism. In the context of contemporary medical imaging (multi)fractal techniques have only been leveraged for static feature extraction for the purposes of classification using pre-trained \ac{DNN}s \cite{mohammed2018neural, roberto2021fractal, xiong2023spatial, Maria2023}.  

Fully end-to-end approaches have only been recently developed in the computer vision community. For texture classification, Yong Xu et al. \cite{xu2021encoding} adapted their hand-crafted approach \cite{xu2009viewpoint} to be supported end-to-end. Their work was motivated to complement \ac{GAP} with relaxed soft version of the \ac{MFS}. After computing a soft-histogram over $Q$ level sets, they compute a quantity proportional to the fractal dimension by max-pooling operations of increasing size for each level set $Q$ . They fuse the soft \ac{MFS} with \ac{GAP} using bilinear pooling. 
Chen et al. \cite{chen2021deep} assume that a \ac{CNN} is roughly a counterpart to the wavelet transform and consider that the fractal dimension is encoded throughout the layers and channels of a \ac{CNN}, i.e., it as a function of a four-dimensional tensor (height, width, channel, and layer). They relax the \ac{DBC} algorithm \cite{sarkar1994efficient} to be used during training of the \ac{CNN}. The surface tensor is constructed by imposing size-normalization constraints over spatial and channel dimensions of each layer. The proposed \ac{DBC} algorithm is computed in overlapping windows and they learn a set of soft histograms for each $l$. Like~\cite{xu2021encoding} they also use a \ac{GAP} layer on the embedding, but instead of bilinear pooling, they simply concatenate both feature maps. 

For the task of few-shot learning, \cite{zhou2023fractal} introduced the fractal dimension as a prior for intra (inter) class similarity (dissimilarity), under the pretense that image regularity will differ across semantic categories. The \ac{DBC} is also adapted, this time solely by removing the non-differentiable rounding functions. Since they assume a perfect linear relationship, they skip the \ac{OLS} step required to estimate the scaling exponents \cite{xu2009viewpoint} and collect the point-wise estimate of the \ac{DBC} in order to eventually computen the so-called \textit{fractal embedding}. 

Recently, \cite{chen2024mfen} proposed the Multi-layer Fractal Encoding Network (MFEN). 
The box-dimension is computed by exhaustive thresholding of the $\ell_2$-norm distance of all points. The fractal dimension of each layer is derived using the ridged regression method. 
These fractal features are concatenated with the per-layer global average pooling output and a final dense \ac{MLP} outputs the final predictions. 
\subsection{Statistical Channel Attention Functions}
\label{sec:rel_work:channel_attention}

Squeeze-and-Excitation Networks \cite{hu2018squeeze} (SE) had a profound impact in modern deep learning architectures. SE is a simple \textit{plug in} component that allowed each enconder in a \ac{CNN} to have global receptive field. This was made possible since the input at depth $l$ is recalibrated according to some \textit{channel attention function} $g$. In this case, $g$ is the output of an \ac{MLP} over the \ac{GAP} of some encoder at depth $l-1$. One of the main limitations of the original SE module is the fact that \ac{GAP} is a very limited feature descriptor by virtue of (a) being a global descriptor, by (b) being the simplest statistic one can compute to describe each filter, and (c) by the representation power and compute requirements of $g$. 
The reader is referred to \cite{guo2022attention} where a very comprehensive survey of extensions to SE is presented. 

In \cite{gao2019global}, the second-order statistics are modelled through the covariance matrix across channels. However, they had to reduce the total number of channels in layer $l$, $C_l$, by means of $1 \times 1$ convolution to mitigate the introduced computational burden. The authors of \cite{lee2019srm} propose the style-based recalibration (SRM) module that uses both global average and standard deviation pooling to expand the receptive field of $f_l$. They learn a non-linear map of the mean and variance of each channel using channel-wise dense layers. In~\cite{qin2021fcanet}, the frequency channel attention (FCA) module was introduced. The authors illustrated how \ac{GAP} is proportional to the lowest frequency component of the 2-dimensional Discrete Cosine Transform (DCT). 

Specifically to Medical Imaging Segmentation, \cite{roy2018recalibrating} propose recalibration using two branches: the \textit{channel squeeze spatial excite} branch (cSE), a direct implementation of the original SE module, and the \textit{spatial squeeze channel excite} (sSE) branch, that acts as spatial attention \cite{oktay2018attention}. The output layer is an element-wise max-out between these two branches (scSE). 
The pursuit of channel-wise descriptions was further explored in \cite{pereira2019adaptive, rickmann2020recalibrating}, but explicitly tailored to 3D segmentation and without emphasis in points (a) and (b), but in terms of (c). Related to the latter, other approaches demand an increasing amount of compute and/or architectural changes, such as introducing Transformers \cite{wang2022uctransnet, huang2022scaleformer, azad2022medical}.
\section{Preliminaries}
\label{sec:methods:perliminaries}
In the remainder of this work, we will denote the input of a \ac{DNN} encoder, $\Psi$,  with an input tensor $\boldsymbol{X} \in \mathbb{I}_0$, so that $\Psi_l(\boldsymbol{X} )= \Psi_l  \circ \ldots \circ \Psi_2 \circ \Psi_1(\boldsymbol{X})$ where $\Psi_l: \mathbb{I}_{l-1} \rightarrow \mathbb{I}_l$ for $\mathbb{I}_l \equiv \mathbb{R}^{H_l \times W_l \times C_l}$. Thus, $l \in \{1,\ldots, L\}$ indexes the layers of $\Psi$. 
Furthermore, $\Psi_{lhwc}$ signifies the $c$-th channel, at position $(h,w)$ of the $l$-th layer. Assume henceforth that this indexing  is consistent throughout and that index omission implies tensor slicing, e.g. $\Psi_{lc} \equiv \Psi_{l::c}$.

Our contributions follow closely an early treatise on multifractal measures presented in \cite{evertsz1992multifractal}. Although uncesserily complex for the purposes of the paper, we refer the reader to \cite{abry2015bridge, jaffard2019multifractal, wendt2022multifractal} for more recent and complete formalisms.
\vspace{5mm}

\textit{Fractal Geometry} is concerned with the study of sets that have fine-structure, i.e., detail at all scales \cite{falconer2004fractal}. Often these sets are also (either exactly or statically) \textit{self-similar}, which is to say that small portions of the set may be magnified and distorted in such a way as to resemble the set as a whole. Typically, one can generate such sets by means of simple, iterative procedures \cite{falconer2004fractal}.

A self-similar set $\mathbb{I}\subseteq \mathbb{R}$ can thus be partitioned into $N$ equal parts scaled by a factor $r=\frac{1}{N^{-D}}$. Is is illustrative to consider the following identity:
\begin{equation*}
	N \cdot \left(\frac{1}{N^{-D}}\right)^D = N \cdot r^D = 1, 
\end{equation*}
where $N>0, D \in \mathbb{R}_{0}^+$. This in turn implies that:
\begin{equation}
\label{eq:hausdorff}
	N = r^{-D} \iff D = - \frac{\log{N}}{\log{r}}. 
 \end{equation}

For these types of sets, $D$ is precisely the Hausdorff-Besicovitch dimension \cite{falconer2004fractal}. Crucially, a set is a fractal if its Hausdorff-Besicovitch dimension exceeds its topological dimension (i.e., its Lebesgue covering dimension). 

In this manuscript, we will not be concerned with the strict, general definition of the Hausdorff-Besicovitch dimension. Note only that that its computation is involved even for very simple sets \cite{salat2017multifractal}. We are however interested in good practical heuristics that either approximate it, or are at the very least proportional to it. 
We thus consider satisfactory that \eqref{eq:hausdorff} has been found to serve as a good empirical prior in general \cite{falconer2004fractal}. Onwards, assume that a \textit{fractal} set $\mathbb{F}$ is any set for which the relationship in~\eqref{eq:hausdorff} can be verified at least  as we count the number of sets of size $r$ that cover $\mathbb{F}$, $N_r(\mathbb{F})$, as $r\rightarrow 0$. In practice, we will compute a \textit{statistical} estimate of the dimension $\hat{D}(\mathbb{F})$ as:
\begin{equation}
    \label{eq:hausdorff_approx}
    \hat{D}(\mathbb{F}) \approx \text{slope of }(\log N_r(\mathbb{F}), -\log r)_{r \in \mathcal{R}},
\end{equation}
where $\mathcal{R}$ is a finite set of discrete $r$.  $\hat{D}(\mathbb{F})$ is the so called \textit{box dimension} of $\mathbb{F}$ \cite{karperien2016multifractal}. Onwards, we will use the terms fractal and box dimension interchangeably and will refer to the theoretical Hausdorff-Besicovitch dimension explicitly if there are meaningful theoretical differences. 

\subsection{Multifractal formalism}
\label{sec:background:multi}
\begin{figure}
    \centering
    \includegraphics[width=1.00\columnwidth]{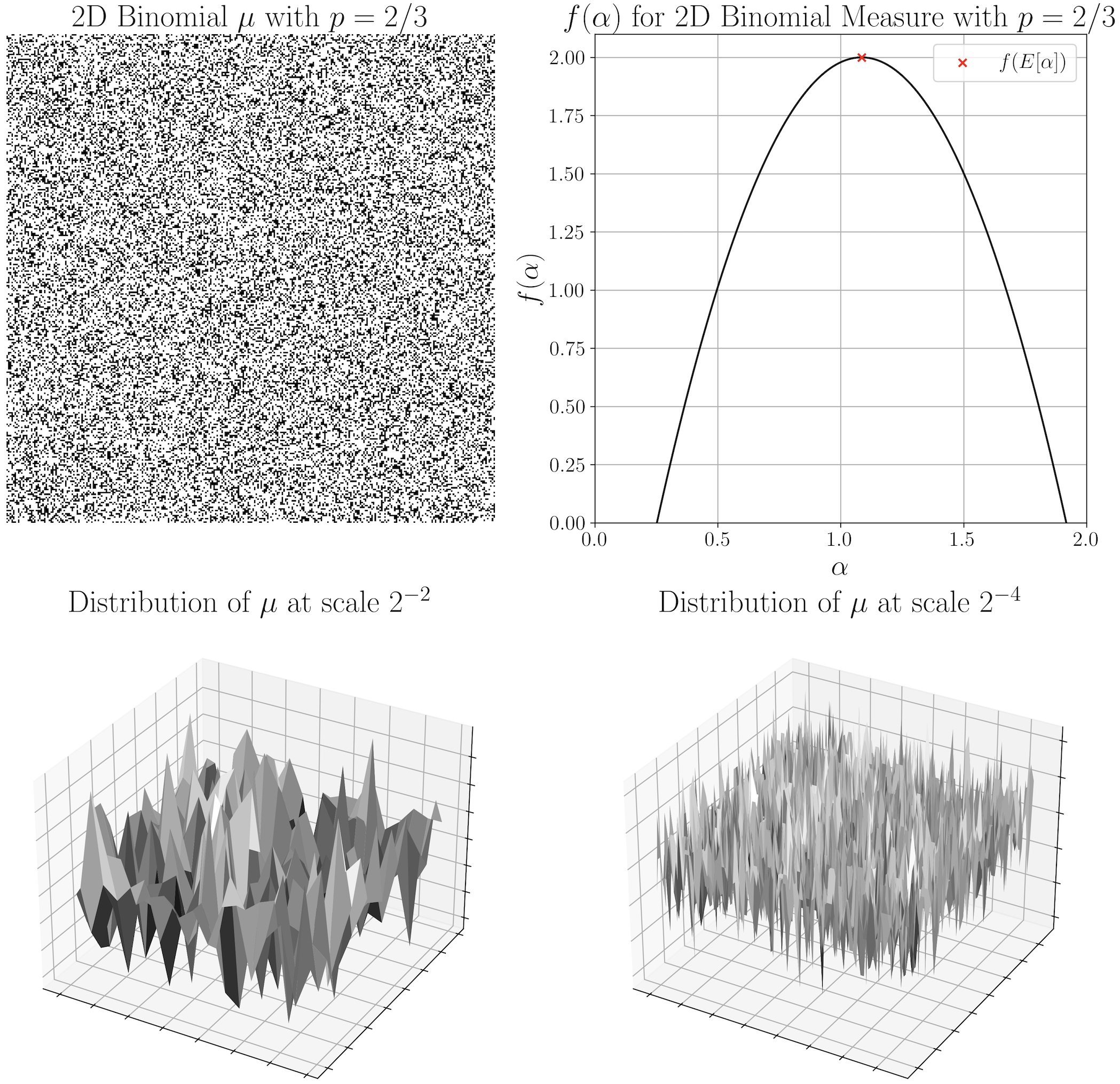}
    \caption{(Top-left) The realization of binomial measure $\mu$ with $p=2/3$ supported in 2-dimension Euclidean space. (Top-right) The associated multifractal spectrum. Notice that $f$ is a concave parabola and $\mu$ is singular, but $f(E_{x}[\alpha(x)])$ will still be associated with the dimension of the support of $\mu$ which is 2. (Bottom-left+right) 3D visualization of distribution of the $\mu$ at two different scales $2^{-k}$, $k=\{2,4\}$. Notice how the surface of $\mu$ appears to be more irregular as $k$ increases.}
    \label{fig:binomial}
\end{figure}

In essence, Multifractal formalism extends Fractal Geometry from sets to measures \cite{evertsz1992multifractal}. These measures may be supported on some Euclidean or fractal set. 

Consider a measurable space $(\mathbb{I}, \mathcal{B}(\mathbb{I}), \mu)$, where $\mathcal{B}$ is a Borel $\sigma$-algebra. 
Should the measure $\mu$ be self-similar, its value in an open ball of radius $r$, $B_r\in \mathcal{B}(\mathbb{I})$, is expected to \textit{roughly} have a power relationship with $\alpha$, i.e., $B_r\sim r^\alpha$. However, now we are envisioning that more than one scaling exponent may govern $\mu$, so it is instructive to construct a self-similar measure and observe the cases where the term \textit{multifractal} becomes necessary. Specifically, we will focus on the multinomial measures.

Suppose that $\mathbb{I} = [0, 1]$, but $\mu$ will assign mass according to some parameter $p \in [0,1]$ at $\mathbb{I}_0 := [0, 2^{-1})$ (left) and $(1-p)$ at $\mathbb{I}_1:=[2^{-1}, 1)$ (right). We will extend this inductively and construct a binomial expansion $\mathbf{w}_k$ for each dyadic interval by marking the left and right expansions with $0$ and $1$, respectively. 
Inductively, let $\mathbb{I}_{{k+1}} = \mathbb{I}_{{k0}}\cup\mathbb{I}_{{{k}1}}$, and collect all possible expansions on $\mathbf{w}_{k} =(w_1, w_2, \ldots, w_k)$, $w_i \in \{0, 1\}$. Then at \emph{scale} $k$, the measure will be defined in the following collection $\mathbb{I}_{\mathbf{w}_{k}}:= \cup_i [i2^{-k}, (i+1)2^{-k})$, where $\mathbf{w}_{k}=(w_1, w_2, \ldots, w_i, \ldots w_{2^k})$, $w_i \in \{0, 1\}$. Note that we choose by convention the expansion with infinitely many zeroes to account for dyadic points (i.e., all $x=\frac{m}{2^{-k}}$, where $m$ and $k$ are integers). Then, by construction:

\begin{equation}
    \mu(\mathbb{I}_{\mathbf{w}_{k}}) = p^{n_0(\mathbb{I}_{\mathbf{w}_{k}})}(1-p)^{k - n_0(\mathbb{I}_{\mathbf{w}_{k}})},
    \label{eq:binomial_measure}
\end{equation}
where $n_0(\mathbb{I}_{\mathbf{w}_{k}})$ denotes the number of zeroes in $\mathbf{w}_{k}$. Note that $\mathbb{I}_{\mathbf{w}_{k}}\in \mathcal{B}(\mathbb{I})$, for any  $\mathbf{w}_k$.

Formally, if for some function $g$ evaluated at arbitrary $x$ there exists $r, x_0$ and $C$ so that:
\begin{equation}
    \alpha(x) = \sup_{|x - x_0| \leq r} \{\alpha': |g(x) - g(x_0)| \leq C|x-x_0|^{\alpha'}\},
    \label{eq:local_holder}
\end{equation}
then $\alpha(x)$ is \textit{Hölder exponent} or \textit{singularity strength} of $\mu$ at scale $r$ around $x$ \cite{chhabra1989direct, evertsz1992multifractal}. 
Hölder regularity is also often expressed as the distance of $f(x)$ to a polynomial $P(x-x_0)$ with degree $k<\alpha$ \cite{jaffard2019multifractal}. Intuitively, having well-defined bounded derivatives up to order $\alpha$ means that the function tends to vary more smoothly, whereas we expect functions with smaller values of $\alpha$ (especially for $\alpha<1$) to behave more irregularly. It also means that the function will behave in a scale-free fashion around $x$ with exponent $\alpha(x)$. 

For our purposes, $\alpha_r$ is the value \eqref{eq:holder_log} within a $r$-neighbourhood around $x$. This neighbourhood is a set $\mathbb{I}_r\in \mathcal{B}(\mathbb{I})$ and we assume that $x \in \mathbb{I}_r \implies \alpha_r(x)\approx \alpha(\mathbb{I}_r)$. If $\mu$ is self-similar at $x$, then we expect to derive $\alpha(x)$ by observing $(\alpha_r(x), x)$ for increasingly small discrete $r$, and estimate this quantity analogously to \eqref{eq:hausdorff_approx}, which is commonly demoninated as
 \textit{local Hölder exponent} or \textit{singularity strength} \cite{evertsz1992multifractal, halsey1986fractal, xu2009viewpoint}: 
\begin{equation}
    \alpha_k(x) \sim \frac{\log \mu(\mathbb{I}_{\mathbf{w}_{k}})}{\log 2^{-k}}.
    \label{eq:holder_log}
\end{equation}
The value of $\mu(\mathbb{I}_{\mathbf{w}_{k}})$ depends solely on $p$ and $n_0(\mathbb{I}_{\mathbf{w}_{k}})$. In practice, the coarse local Hölder exponent at $x$
is computed by evaluating \eqref{eq:holder_log} at several \emph{scales} $k$ and an estimate estimate is produced analogously to \eqref{eq:hausdorff_approx}. 

 The concept of a multifractal measure becomes clear when one observes that $\mu$ is exactly the \textit{Binomial} measure.  Specifically, that the number of points that scale with $\alpha$, $N_k(\alpha)$, can be attained by computing the binomial expansion, and thus: 

\begin{equation}
    N_k(\alpha)=\binom{k}{n_0(\alpha_k)k} \sim (2^{-k})^{-f(\alpha_k)},
    \label{eq:binomial_approx}
\end{equation}
so that $f(\alpha)=-\log_2(n_0(\alpha)^{n_0(\alpha)}(1 - n_0(\alpha)^{1-n_0(\alpha)})$, by leveraging 
Striling's approximation \cite{evertsz1992multifractal}. This relation $N_k(\alpha)\sim r^{-f(\alpha_k)}$, is homologous to \eqref{eq:hausdorff}. For measures where such a relation holds, the value of $f(\alpha)$ will exactly be the Hausdorff-Besicovitch dimension of the following level set~\cite{evertsz1992multifractal,xu2009viewpoint,xu2021encoding}:
\begin{equation}
    \mathbb{I}^{\alpha} = \{x \in \mathbb{I}: \alpha(x)=\alpha\}.
    \label{eq:level_set}
\end{equation}
Since $f$ maps $\alpha$ to a spectrum of fractal dimensions of $\mathbb{I}^{\alpha}$, it is henceforth called the \emph{multifractal spectrum} (\ac{MFS}).

These relations can be derived for the more general Multinomial measures parameterized by $m$ parameters so that $\mu(\mathbb{I}_{\mathbf{w}_{k}}) = \prod_{i=1}^{m}p_i^{n_i(\mathbb{I}_{\mathbf{w}_{k}})}$ \cite{evertsz1992multifractal}. There are also other ways of computing $f$ by using Wavelet transforms or by \ac{DFA}, and we recommend \cite{salat2017multifractal} for an in-depth overview of these methods. 

Our contribution relates more closesly with the \textit{Histogram Method} given that we are focusing on imaging modalities. Empirical results appear to favor the  method in this context of Computer Vision, especially if one computes several different measures (i.e., filters) over the same \textit{static} phenomena (i.e., image) \cite{vehel1994multifractal, xu2009viewpoint, hafner2015local}. 
There are also other ways of computing $f$ by using Wavelet transforms or by \ac{DFA}, and we recommend \cite{salat2017multifractal} for an in-depth overview of these methods

The \textit{Method of Moments} is arguably more adopted in general. It uses the \textit{Legendre} transform to write $f$ and $\alpha$ as functions of the normalized moments $q$ of $\mu$, $\tau(q)$:  
\begin{equation}
    f(\alpha(q))=q \alpha(q) - \tau(q) \text{ where } \alpha(q) = \frac{\partial \tau(q)}{\partial q}, q \in \mathbb{R}.
    \label{eq:moments}
\end{equation}
It is also pivotal for defining the \textit{set of generalized dimensions} $D_q$, and the interested reader is referred to \cite{hentschel1983infinite, karperien2016multifractal} for additional details. For this paper, we mainly use the link between $D_q$ and $f(\alpha)$ as a convenient formal construct for our proofs.

\subsection{Squeeze-and-Excite \cite{hu2018squeeze}}

Squeeze and excitation networks (SE) were originally studied by \cite{roy2018recalibrating}, and we will onward use the the acronym cSE (channel SE) for consistency with medical image segmentation literature, even though cSE and SE are functionally equivalent.  

The output of each encoder layer $\Psi_l$ is characterized by a ``squeeze'' function $g:\mathbb{I}_l \rightarrow \mathbb{R}^{C_l}$ such that:
\begin{equation}
    g(\Psi_l(\mathbf{X})) =  \sigma(\mathbf{W_2}\operatorname{ReLU}(\mathbf{W}_1\text{GAP}(\Psi_l(\mathbf{X})))),
    \label{eq:se}
\end{equation}
where GAP denotes spatial global average pooling, $\mathbf{W}_1 \in \mathbb{R}^{\lfloor\frac{C_l}{s^*}\rfloor \times C_l}$, $\mathbf{W}_2 \in \mathbb{R}^{C_l \times \lfloor\frac{C_l}{s^*}\rfloor}$, for some $1\leq s^* < C_l$, and $\sigma$ is the sigmoid activation function. The output of the encoder at level $l$ for $l < L-1$ thus becomes:
\begin{equation}
    \Psi_l^{\text{Excited}}(\mathbf{X}) = \Psi_l(\mathbf{X} ) \odot g(\Psi_l(\mathbf{X} )),
    \label{eq:excite}
\end{equation}
where $\odot$ is the (broadcastable) element-wise product. 

\subsection{Spatial and Channel SE\cite{roy2018recalibrating }}
For the purposes of semantic segmentation of medical images, it is proposed to perform the squeeze operator along the spatial dimensions, in order to have an attention function with the original feature map resolution.

The authors propose an additional branch, where the pooling operator of \eqref{eq:se} is replaced by a learnable $1\times1$ convolution layer, producing $g^{\text{Spatial}}(\Psi_l(\mathbf{X}))$. The authors propose the final activations to be 
\begin{equation}
    \Psi_l(\mathbf{X}) = \max(\Psi_l(\mathbf{X}) \odot g(\Psi_l(\mathbf{X}),\Psi_l(\mathbf{X}) \odot g^{\text{Spatial}}(\Psi_l(\mathbf{X})),
    \label{eq:sc-se}
\end{equation}
where $\max$ is performed element-wise, i.e., the function is a maxout layer \cite{goodfellow2013maxout}.
\subsection{Style-based Recalibration \cite{lee2019srm}}
Style-based Recalibration (SRM) integrates the standard deviation as a proxy for style in the context of style-transfer. Denoting a global standard-deviation pooling layer as GSP, the squeeze function is defined as: 
\begin{equation}
    g(\Psi_l(\mathbf{X})) = \phi([\text{GAP}((\Psi_l(\mathbf{X}))),\text{GSP}((\Psi_l(\mathbf{X})))]),
    \label{eq:srm}
\end{equation}
where $\phi:\mathbb{R}^{C_l \times 2} \rightarrow \mathbb{R}^{C_l}$ is a learnable linear map. An additional batch-normalization layer is applied to $g$.

\subsection{Frequency Channel Attention \cite{qin2021fcanet}}
Frequency Channel Attention networks (FCA) were designed to generalize the ``squeeze'' operation in SE. 

The authors observe that global average pooling acting on feature maps of fixed spatial dimensions is a special case of of the Discrete Cosine Transform (DCT). More precisely, it is equal to the lowest frequency component up to a fixed constant factor \cite{qin2021fcanet}. FCA thus enables to use the entirety of the DCT transform to describe feature maps by partitioning each layer in $k$ groups, whereupon action of a distinct DCT basis is estimated. 

Specifically, the $C_l$ channels of each layer $l$ in are split into $k$ groups such that $[\Psi_l(\mathbf{X})_1, \ldots, \Psi_l(\mathbf{X})_k]=\Psi_l(\mathbf{X})$. 
The two-dimensional DCT filters are pre-computed and stored on a tensor $\mathbf{B}$: 
\begin{equation}
    \mathbf{B}_{l,h,w}^{i,j}=\cos \left(\frac{\pi h}{H_l}\left(i + \frac{1}{2}\right)\right) \cos \left(\frac{\pi w}{W_l}\left(j + \frac{1}{2}\right)\right).
    \label{eq:cos}
\end{equation}

The squeeze operation is performed for each of the $k$ groups:
\begin{equation}
    g(\Psi_l(\mathbf{X}))_k = \sum_h \sum_w \Psi_l(\mathbf{X})_kB_{l,h,w} ^{i_k,j_k}, 
    \label{eq:fca:one}
\end{equation}
$i_k,j_k$ depend on $k$ since ensuring  a distinct DCT basis for each of the $k$ groups $ g(\Psi_l(\mathbf{X})) := \sigma(\mathbf{W}[g(\Psi_l(\mathbf{X}))_1, \ldots, g(\Psi_l(\mathbf{X}))_k])$, so that $\mathbf{W} \in \mathbb{R}^{C_l \times C_l}$.


\section{Methods}
\label{sec:methods}
We look at the $l$-th output of encoder head $\Psi_l = [\mu_{l{c_1}}, \ldots, \mu_{l{C_l}}]$, $\mu_{lc} \geq 0$, 
i.e., a tensor of unnormalized measures (filters) that we assume are meaningful for the downstream task. We adjust the responses of the encoder using \textit{learnable functions} of the local (multifractal) or global (monofractal) interactions between the singularity strengths derived across each channel/measure of $f_l$. We start by demonstrating that the proposed recalibration functions  indeed relate to quantities that relate to $f(\alpha)$.

Our proofs will consider measures $\nu = (\mu_q)_{q=1}^Q$ that take the multinomial form. By construction, each $\mu_q$ is a binomial measure with potentially unique support and we model and let $p^{(q)}(\alpha^{(q)})$ be the normalized density of each exponent in $\nu$ so that:
\begin{equation}
    \label{eq:multinomial_pk}
    p(\alpha_\nu) := [p^{(1)}(\alpha^{(1)}) / Z, \ldots, p^{(Q)}(\alpha^{(Q)}) / Z],
 \end{equation}
where $Z = \sum_{q} p^{(q)}(\alpha^{(q)})$. Note however that these multifractal measures have enjoyed great success in practice \cite{karperien2016multifractal, hafner2015local}. We also introduce learnable parameters in their implementation, thus allowing to characterize the scaling nature of other multifractal phenomena beyond multinomial cascades.
\subsection{Differentiable scaling exponent computation}
As mentioned in Section \ref{sec:background:multi}, the Hölder exponent can be computed in an analogous fashion to the fractal dimension \eqref{eq:hausdorff_approx}. Defining a finite scale set $\mathcal{R}=\{k_1,\ldots, k_r: i < j \implies k_i < k_j \}$, we can derive the following dyadic cubes $ \left[ i_h - 2^{-k}, i_h + 2^{-k} \right) \times \left[ i_w - 2^{-k}, i_w + 2^{-k} \right)$. For a cube $B_k(x)$ centered on $x$ we can approximate the local singularity strength for the $c$-th channel of the $l$-th layer as the power-law relationship manifested at finitely many resolutions $k$. We approximate the local Hölder \eqref{eq:holder_log} exponents by solving:
\begin{equation*}
    \alpha(x)\approx \text{slope of }(\log \mu (B_k(x)) , \log k)_{k \in \mathcal{R}},
\end{equation*}
for all $x$. Concretely, we attain this value through the ordinary least squares solution:
   \begin{equation}
       \alpha(x)=\frac{
           \sum_{k}\left(\log\mu(B_k(x))  - \frac{1}{|\mathcal{R}|}\sum_{k'}\log (\mu(B_{k'}(x))\right)
}{
    \sum_{k}\left(\log k - \frac{1}{|\mathcal{R}|}\sum_{k'}\log k'\right).
}
    \label{eq:ols}
\end{equation}
In our implementation, $\mu(B_k(x))$ is computed by convolving each $\Psi_{lc}$ with a kernel of ones of size $k$ and stride 1 (a static depth-wise convolution layer in practice). This operation extends trivially to tensor operations, so $\alpha_{lc}(x)$ can be efficiently computed by perfoming this operation in a depth-wise convolution layer, readily available in modern deep learning frameworks \footnote{E.g., PyTorch through \texttt{torch.nn.Conv2d} (with \texttt{groups=in\_channels}) or TensorFlow's \texttt{DepthWiseConv2D}.}.
\subsection{Fractal-based Recalibration}
\label{sec:methods:mono}

Recalling the notation set on the Section \ref{sec:methods:perliminaries}, suppose we collect each Hölder exponent in:
\begin{equation}
     \boldsymbol{\alpha} := \left( \alpha_k(x) \right)_{x \in \operatorname{supp}\mu}
    \label{eq:holder_matrix}
\end{equation}
and let ${\alpha}_k(\cdot)$ be an approximation of \eqref{eq:ols} using resolutions $\mathcal{R}=\{1, \ldots, 2^{-k}\}$.

\begin{theorem}[Fractal recalibration]
    Supose that $\mu$ is a multinomial measure as in equation \eqref{eq:multinomial_pk} with multifractal spectrum $f$, then almost surely (a.s.) for $\mathcal{R}=\{1, \ldots, k\}$ for sufficiently large $k$:
    $$ \mathbb{E}_x [\boldsymbol{\alpha}] \propto  D(\operatorname{supp}\mu),$$ 
    \label{proof:mono}
    where $D$ is the box-dimenson of the support.
\end{theorem}

\textit{Proof}: We prove the non-trivial case for non-singular multinomial measures using an argument similar to Evertsz and Mandelbrot \cite[Section 2.4]{evertsz1992multifractal}. By definition $\mu(\mathbb{I}) = 1$ and so $\mu(\mathbb{I}^\alpha) \leq \mu(\mathbb{I})$.
Note also that $\mu(\mathbb{I}^\alpha)= r^{-f(\alpha)}r^\alpha\leq 1$ (refer to \eqref{eq:binomial_approx}), and thus it must be that $f(\alpha) \leq \alpha$. 
For multinomial measures, $f(\alpha)$ is concave. Also for this case, and referring back to the Legendre transform in \eqref{eq:moments}, we can assume $q \geq 0$, so the exponent $\alpha$ that maximizes $f$ needs to verify $\frac{\partial f(\alpha(q))}{\partial \alpha }=0$ --- this extremum is a maximum if and only if $q=0$, which is equivalent box dimension of the support of the measure \cite{karperien2016multifractal}. Referring back to the shape of $f$ (see Eq. \eqref{eq:binomial_approx} or Figure \ref{fig:binomial}), clearly $\mathbb{E}[\alpha]$ maximizes $f(\alpha)$. 

In order to show that the expectation over $x$ relates to this quantity, we again extend the arguments of Evertsz and Mandelbrot in \cite[Section 4.2]{evertsz1992multifractal}: leveraging the law of large numbers, for any $x \in \operatorname{supp}(x)$:
$$ {\alpha}(x)\overset{\text{a.s.}}{=}\mathbb{E}(\alpha), \text{ where }  \alpha \text{ in the r.h.s. is in the domain of the MFS of $\mu$: } (\alpha, f(\alpha))$$
 Denoting $|\operatorname{supp}\mu| = n \in \mathbb{R}^+$, it follows that $
\mathbb{E}_x[\boldsymbol{\alpha}] \overset{\text{a.s.}}{=} n^{-1} D(\operatorname{supp}\mu)$
\qedwhite\\

\subsubsection{Monofractal recalibration}
This theorem provides a fundamental explication to empirical observations: estimations of $\alpha$  tend not to vary meaningfully in the support of the measure \cite{karperien2016multifractal, salat2017multifractal}. 
\begin{remark}[Monofractal recalibration]
    If $\mu$ is uniform and supported on fractal set $\mathbb{F}$, $\mu$ is said to be monofractal, i.e., it is governed by a single scaling exponent. Under these conditions, the multinomial measure will reduce to the Lebesgue measure $\lambda$. However, $\mathbb{E}[\alpha(x)] \propto {D}(\mathbb{F})$ as in \eqref{eq:hausdorff}, even though $\lambda(\mathbb{F})=0$.
    \label{proof:mono2}
\end{remark}
\textit{Proof}: The probability mass is uniform so $\mu(\mathbb{I}^{\alpha_i})=\mu(\mathbb{I}^{\alpha_j})$ for any $i,j$, i.e., the mass is evenly distributed with respect to the terms of the the $m$-ary expansion of any $x$, and so $\mathbb{I}=\{\alpha: x \in \operatorname{supp}\mu\}$.  
The \ac{MFS} thus collapses to a single value $\alpha$, which can be recovered using the expectation operator in Theorem \ref{proof:mono}.
\qedwhite
\subsubsection{Multifractal recalibration}
\label{sec:methods:multi}
The previous recalibration strategy is suitable for when each (learned) measure $\mu$ is well characterized by a monofractal . Notwithstanding, it does not characterize the entirity of the \ac{MFS}, it can only capture a value associated with the dimension of its support.

    Remember from Section \ref{sec:background:multi} that $f(\alpha)$ is the \textit{Haurdorff-Besicovich} dimension of $\mathbb{I}^\alpha$. We know that $\mu(\mathbb{I}^\alpha)=N_k(\alpha)(2^{-k})  \sim 2^{-f(\alpha) + \alpha}$. If one treats the Hölder exponent of a binomial cascade at the $k$-th scale $\alpha$ as a random variable, then by the Gaussian Central Limit theorem, for sufficiently large $k$, the following two statements hold \cite{evertsz1992multifractal}:

\begin{equation}
    p_k(\alpha) \propto \exp\left\{\frac{1}{2}\left(\frac{\alpha - \alpha_0}{\sigma / \sqrt{k}}\right)^2\right\}
    \label{eq:gaussian_clt}
\end{equation} 
aroud most probable scaling exponent $\alpha_0$.
Recall that  $\mathbb{E}_x [\boldsymbol{\alpha}] \rightarrow \alpha_0$ as $k\rightarrow \infty$.
Should $\mu$ be supported on a set with Hausdorff-Besicovich dimension $D$ \cite{evertsz1992multifractal}:
\begin{equation}
f_k(\alpha) \approx D + \frac{1}{k} p_k(\alpha).
    \label{eq:f_prob}
\end{equation}
Hence, for our purposes, it is fair to assume that the support has fixed dimension, so our inductive bias can be on $p_k$, since $p_k \propto f_k$. 

It is also very possible that some phenomena may not be aptly described solely by this quantity. Indeed the conditional likelihood of the singularities may vary depending on the context. We set forth \emph{Multifractal Recalibration} to address this issue, assuming there the quantity we want to model may also depend on the context (i.e., the desirable mapping $\boldsymbol{X} \mapsto \boldsymbol{Y}$).

\begin{definition}[Data-dependent multinomial measure]
    Suppose we build $\mu^*$ by weighing multinomial $\mu$ as a function of a labeling function $l: \mathbb{I} \rightarrow \{0, 1\}$  defined as the restriction $\mathbb{Y}=\{x \in \mathbb{I}: l(x) = 1 \}$. Then, the data-dependent multinomial measure is given by:
    $$ \mu^* (\mathbb{I}) = \frac{\mu ( \mathbb{I} \cap \mathbb{Y})}{\mu (\mathbb{Y})} $$ 
       \label{def:data_measure}
\end{definition}
Note that $\mu^*$ has the same scaling behaviour as $\mu$, but it assignes all mass to restriction $\mathbb{Y}$. 

%
\begin{remark}[Multifractal recalibation]     Consider all points $x$ in some level set indexed by $q$, i.e., $\mathbb{I}^{\alpha(q)} \subseteq \mathbb{I} = \{x: \alpha(x) = \alpha(q)\}$\footnote{Using index $q$ is merely evocative, but not equivalent to, using the Legendre transform.}. Consider  measures $\mu, \mu^*$ supported on $\mathbb{I}$, the latter being constructed from $\mu$ and some labeling function $l(x)$.  Then any measure $\hat{\mu}$ that verifies: 
    $$\hat{\boldsymbol{\alpha}} = \arg \max_{\boldsymbol{\alpha}'} \frac{1}{Z} \exp (-(\boldsymbol{\alpha}'- \boldsymbol{\alpha}^*)^2), $$
has to have perfect level-set likelihood ratio $\frac{\hat{\mu}(\mathbb{I}^{\alpha(q)})}{\mu^*(\mathbb{I}^{\alpha(q)})} = 1 \, \forall q$, 
where $Z = \int \exp \{-(\boldsymbol{\alpha}'- \boldsymbol{\alpha}^2\} dq$.
    \label{proof:level_set_soft}
\end{remark}

\textit{Proof:}
According to the definition of coarse Hölder exponent \eqref{eq:holder_log}, for any $x, q$ such that $x \in \mathbb{I}^{\alpha(q)}$: 
$$\alpha'_q(x) := \frac{\log\mu'(\mathbb{I}^{\alpha(q)})}{\log 2^{-k}} \text{ and }  \alpha^*_q(x)  := \frac{\log\mu^*(\mathbb{I}^{\alpha(q)})}{\log 2^{-k}}.$$
For sufficiently large fixed $k$, we can disregard the denominators. Moreover, since $Z$ does not depend on $q$, pick arbitrary $\alpha(q)$:
\begin{align*}
     &\ {\alpha}'_q(x)- {\alpha}^*_q(x)  \approx \log \mu'(\mathbb{I}^{\alpha(q)})-\log \mu^*(\mathbb{I}^{\alpha(q)})  \\ 
     &\implies \exp \left\{-\left( \log \mu(\mathbb{I}^{\alpha(q)})-\log \mu^*(\mathbb{I}^{\alpha(q)})\right)^2 \right\} = \exp \left\{-\left(\log\frac{\mu'(\mathbb{I}^{\alpha(q)})}{\mu^*(\mathbb{I}^{\alpha(q)})} \right)^2 \right\}.
\end{align*}

Thus the maximizer $\boldsymbol{\alpha}'$ of the above stated Gaussian function must have {level-set likelihood ratio} $\frac{\mu'(\mathbb{I}^{\alpha(q)})}{\mu^*(\mathbb{I}^{\alpha(q)})} = 1$ for all $q$.
\qedwhite

\subsection{Implementation}
\label{sec:methods:implementation}
Our implementations are grounded on Theorem \ref{proof:mono} and its instantiation in the mono (\ref{proof:mono2}) or multifractal (\ref{proof:level_set_soft}) cases.

We adapt the notation set on \eqref{eq:holder_matrix} 
    to the case where we collect the Hölder exponents captured in the $c$-th filter of the $l$-th layer, given the output of $\Psi_l(\mathbf{X})$:
\begin{equation}
    \mathbf{H}_l \in \mathbb{R}^{H_l \times W_l}:= [\alpha(\Psi_{lc}(x))] \text{,
    for all } x \in \operatorname{supp}\mu.
    \label{eq:average_holder}
\end{equation}
and and suppose we compute $\alpha(\cdot)$ using \eqref{eq:average_holder} with a resolution of $r=2^{-k}$.

The outputs of \eqref{eq:average_holder} are not normalized, which can compromise optimization during gradient descent\cite{chen2024mfen, ioffe2015batch}. Thus, onwards $\mathbf{H}_l$ signifies the empirical singularity exponents after the action on a \emph{distinct} batch-normalization layer for each $l$.

In order to recalibrate the response for the monofractal case we need to determine each filter's  box-dimension. The expectation in \ref{proof:mono} is performed for each channel of $\mathbf{H}_l$, and the global  descriptor is then fed to a squeeze-and-excitation pathway \cite{hu2018squeeze}. The output of $\Psi_l$ will be recalibrated by a function $g:\mathbb{I}_l \rightarrow \mathbb{R}^{C_l}$:
\begin{equation}
    g^{\text{Mono}}(\Psi_l(\boldsymbol{X})) =  \sigma(\mathbf{W_2}\delta(\mathbf{W}_1\text{GAP}(\mathbf{H}_l))),
    \label{eq:ssr}
\end{equation}
$\mathbf{W}_1 \in \mathbb{R}^{\lfloor\frac{C_l}{s^*}\rfloor \times C_l}$, $\mathbf{W}_2 \in \mathbb{R}^{C_l \times \lfloor\frac{C_l}{s^*}\rfloor}$, for some $1\leq s^* < C_l$, and $\sigma, \delta$ are the sigmoid and \ac{ReLU} activation functions, respectively. The output of the encoder at level $l$ for $l < L-1$ thus becomes:
\begin{equation}
    \Psi_l^{\text{Mono}}(\boldsymbol{X}) = \Psi_l(\mathbf{X} ) \odot g^{\text{Mono}}(\Psi_l(\mathbf{X} )).
    \label{eq:ssr_output}
\end{equation}
This response is propagated to $\Psi_{l+1}$, but also to the decoder in case of the U-Net by means of a skip connection.

As mentioned in the previous section, the above recalibration strategy can only capture one scaling exponent per filter. Moreover, it does not account for likely fact that our estimate of $\mu_{lc}$ likely does not meet the conditions of maximization in some downstream task. In other words, we expect our representation $\Psi$ to characterize the entirety of the input $\boldsymbol{X} \sim \mu$ and not necessarily distribute the mass accross each level-sets as a function of our dependant variable of interest $\boldsymbol{Y} \sim \mu^*$. 

To achieve this data-dependant recalibration stated in \ref{proof:level_set_soft}, we allow $Q$ learnable scaling exponents to be learned in gradient descent, so that $\mathbf{H}^*_{l} \in \mathbb{R}^Q$. These exponents induce $Q$ level-sets, so that:
$$p^{(q)}(\alpha(x)) \propto \exp (-s^*_q(\alpha({x}) - \mathbf{H}^*_{lq})^2),$$
where the decay the around the peak of each level-set is modulated by $(s^*_q)^{-1}$, which will vary as the empirical and data-depedant exponents diverge. 

Its general formulation to account for input $\Psi(\mathbf{X})$ is expressed via the following softmax function:
\begin{equation}
    p_{l}^{(q)}(\mathbf{H}_{l}) \propto \frac{\exp(-s^*_q (\mathbf{H}_l - \mathbf{H}_{lq}^*)^2)}{\sum_{q'} \exp(-s^*_{q'} (\mathbf{H}_l - \mathbf{H}^*_{lq'})^2)},       \label{eq:soft_histogram}
\end{equation}
where $q\in \{1, \ldots, Q\}$, and $s_q^* \in \mathbb{R}$, $\mathbf{H}^*_l \in \mathbb{R}^{Q}$  are learnable parameters.  Each layer $\Psi_{l}$ is thus partitioned in $Q$ stochastic level sets, i.e., there is parameter sharing across spatial and channel axes. This expression closely resembles the context-encoder functions leveraged in \cite{zhang2018context, xu2021encoding}. 

The morphology of the \ac{MFS} in this recalibration function is also relaxed due to the data-driven partioning of $\mathbb{I}_l$, which is desirable when the \ac{MFS} morphology differs from a perfect concave parabola \cite{evertsz1992multifractal,chhabra1989direct, halsey1986fractal}. 

\begin{figure}
    \centering
    \includegraphics[width=1.00\columnwidth]{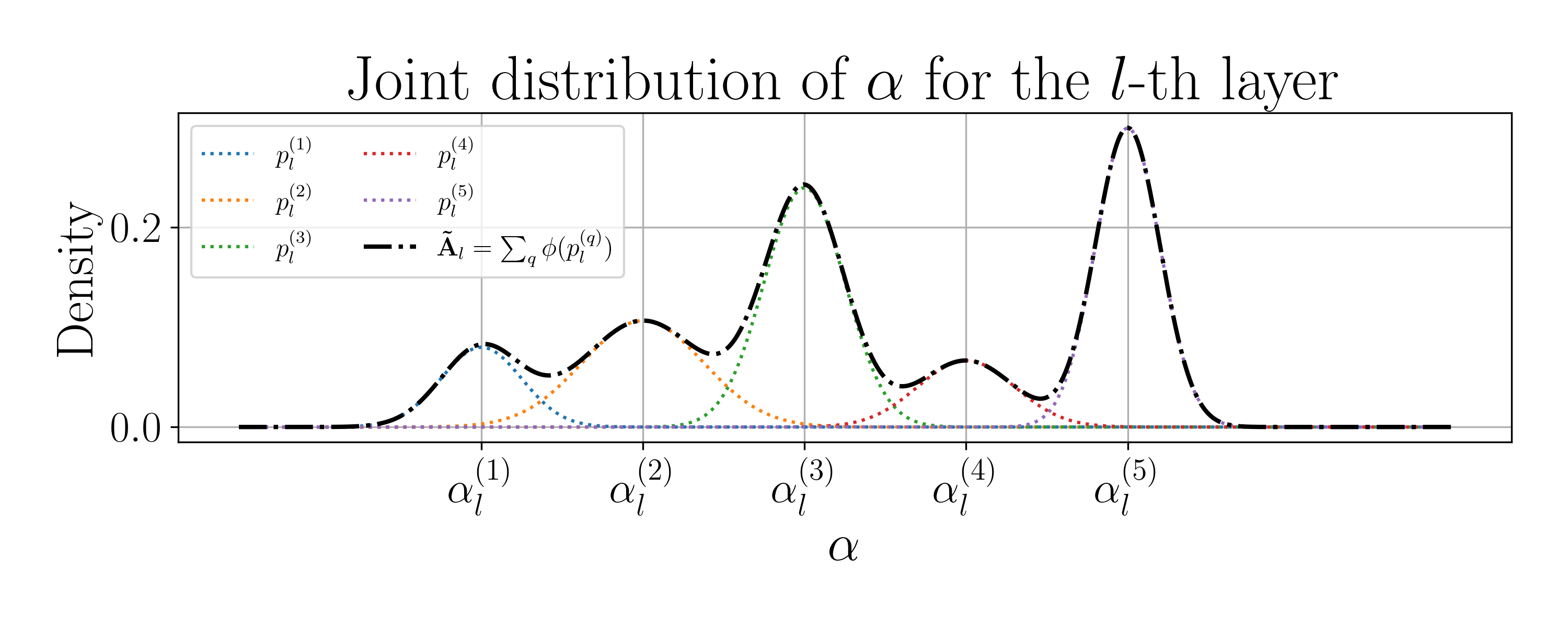}
    \caption{Visualization of the joint encoding of $p_l$ at encoder $\Psi_{l}$. Each level set is characaterized by a Gaussian $p_l^{(q)}$ localized around $\alpha_l^{(q)}$.}
    \label{fig:awesome_figure}
\end{figure}

We surmised that each stochastic level set defined in \eqref{eq:soft_histogram} should be weighted differently depending on the downstream task. We opt for a weighted sum aggregation and learn an orderless pooling function over $q$, a common approach in texture classification \cite{zhang2018context, xu2021encoding, chen2021deep}. We propose to pool the response of $\mathbf{H}_l$ across the $Q$-axis, using a batch-normalization layer $\phi$, so that:
\begin{equation}
\Tilde{\mathbf{H}}_l := g^{\text{Multi}}(\Psi_l(\boldsymbol{X} ))) = \sigma\left(\sum_q \delta(\phi ( p_l^{(q)}(\alpha_{l}^{(q)}) ))\right),
    \label{eq:multi_aggretation}
\end{equation}

 where $\sigma, \delta$ are the sigmoid and \ac{ReLU} activation functions, respectively. Without the activation functions, this aggregation is a mixture of $Q$ Gaussians (as in Figure \ref{fig:awesome_figure}).

 The final step is to fuse this representation with the response of $\Psi_l$. We found that an element-wise sum strategy similar to \cite{zhou2023fractal} to be particularly effective:
 \begin{equation}
    \Psi_l^{\text{Multi}}(\boldsymbol{X}) = \Psi_l(\mathbf{X}) + \Tilde{\mathbf{H}}_l,
    \label{eq:feature_fusion}
 \end{equation}
 where $+$ is the element-wise sum. Note that, by the linearity of convolution, $ \Psi_{lc}^{\text{Multi}} * \mathcal{D}_{lc} = (\Psi_{lc} * \mathcal{F}_{lc}) + (\Tilde{\mathbf{H}}_{lc}* \mathcal{F}_{lc})$, so in this way we ensure that singularity information may be easily disentangled by the decoder filter $\mathcal{D}_{lc}$. 

\section{Experiments}
\label{sec:experiments}
\subsection{Materials}
\label{sec:experiments:mat}
We conducted experiments in three different public datasets from medical imaging modalities that vary significantly in terms of equipment and interventions used. These modalities also present challenging multi-scale characteristics, as well as variations in intensity, lighting, noise artifacts, and texture. Specifically, we selected the ISIC18 (dermoscopy \cite{codella2019skin}), KvasirSEG (endoscopy) \cite{jha2020kvasir}, and BUSI (breast ultrasound)~\cite{busi2020dataset} datasets. As discussed in~\cite{isensee2024nnu}, using three datasets is the median number of recent contributions towards general models for medical image segmentation, provided that the datasets are varied and of sufficiently high quality (i.e., low inter-method and high intra-method variability).
\begin{figure}
    \centering
    \includegraphics[width=1.00\columnwidth]{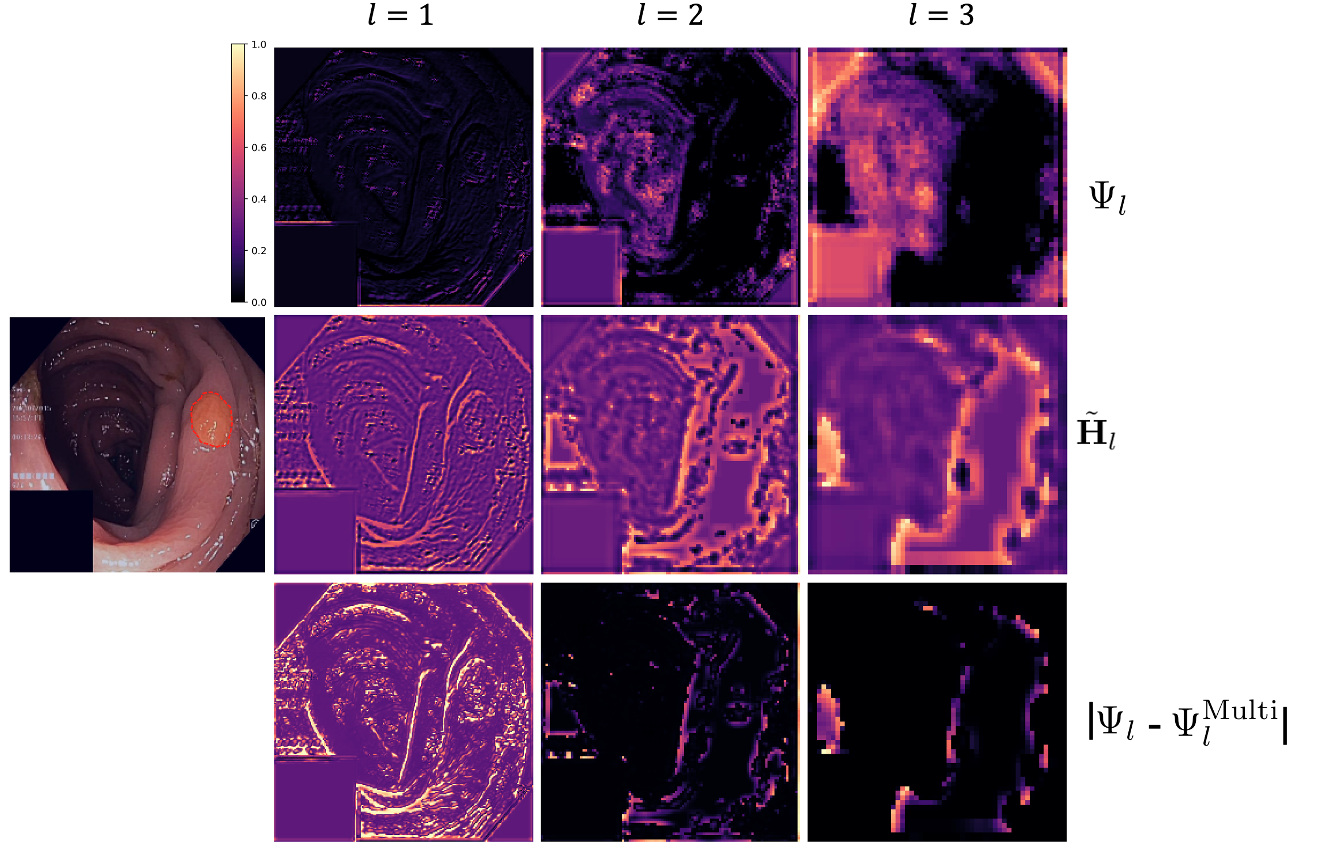}
    \caption{(Left-most) The input KvasirSEG image (from validation set) and target region of interest (highlighted in red). (Right) Layer depth $l \in \{1,2,3\}$ versus normalized response of $\Psi_l$, $\Tilde{\mathbf{H}}_l$, and $|\Psi_l - \Psi^{\text{Multi}}_l|$. 
    Note how $\Tilde{\mathbf{H}}_l$ encodes complementary textural information of $\Psi_l$ for $l=1$. For $l=2,3$ the preferred singularities relate more to luminance changes that highlight anatomical structures. This also illustrates the theory set forth by Vehel at al. \cite{vehel1994multifractal}, where ach level set is associated with a distinct visual primitive.}
    \label{fig:multifractal-dif}
\end{figure}
\subsection{Experimental setup}
\label{sec:experiments:setup}

As prescribed by \cite{isensee2024nnu} we enforce a 5-fold cross-validation setup for all datasets. For each fold, we leave out $10\%$ of each train partition to be used for early stopping. For the case of the BUSI dataset, each split is always stratified over the pathological classification variable since the ratio of benign to malign samples is roughly 2:1 \cite{busi2020dataset}. For all datasets, all image/mask pairs were down-sampled to a $224 \times 224$ spatial resolution using bilinear interpolation and then normalized to have channel values in $[0, 1]$. Standard data augmentation was enforced using random horizontal and vertical flips, each with $p=0.5$. 

We validate our approaches against the original \ac{SE} module \cite{hu2018squeeze}, the concurrent spatial and channel squeeze + channel excitation (scSE) proposed in \cite{roy2018recalibrating}, the \ac{SRM} \cite{lee2019srm}, and the \ac{FCA} \cite{qin2021fcanet} module. For FCA we used the best static configuration comprised of the 16 coefficients associated with the lowest frequencies of the DCT base \cite{qin2021fcanet}. Other than scSE and Multifractal recalibration, we call the other attention function \textit{global} channel attention functions since they do not recalibrate filter responses at the ``pixel level''. 

We use the same baseline as \cite{azad2022medical}, comprised of three encoder/decoder pairs with 32, 64, and 128 channels, and a bottleneck of 256 channels. In cSE, scSE, FCA, and Monofractal recalibration, the \ac{MLP} architecture is the same, and we set $r^*=2$ following \cite{roy2018recalibrating}. For our proposed approaches, we set $\mathcal{R}=\{2, 3, 4\}$ to compute $\mathbf{H}_l$ (see Section \ref{sec:methods}). For Multifractal recalibration we learn $Q=16$ stochastic level sets. 
\begin{figure}
    \begin{center}
        \includegraphics[width=0.95\textwidth]{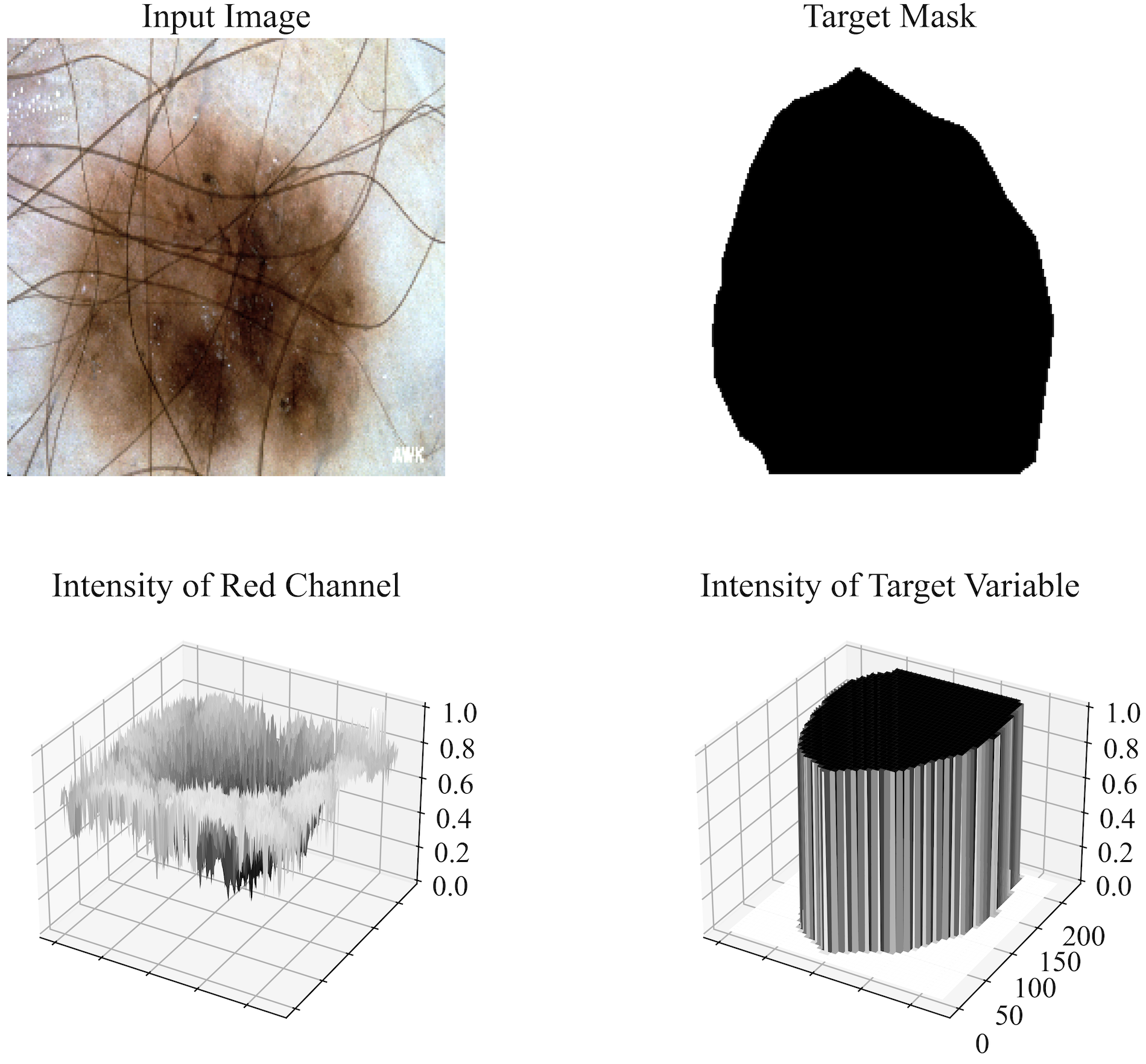}
    \end{center}
    \caption{The intensity distribution of the input is highly irregular. Converly, the target variable is very homogeneous and only relates information of the general region of interest. Sample taken from the ISIC18 dataset.}
    \label{fig:mask_problem}
\end{figure}

Our main research question is whether we can integrate the \textit{multifractality} of $\boldsymbol{X}$ end-to-end to predict $\boldsymbol{Y}$. However, our target variables $\boldsymbol{Y}$ are very regular, even on the borders, likely due to limitations in the annotation procedures (see Fig. \ref{fig:mask_problem}). We contend that, in isolation, the decoder is not explicitly encouraged to approximate an irregular function. With this in mind, and in order to streamline our experimental processes, we study the impact of the recalibration functions solely after each encoder output (before skip-connection and max-pooling as in \cite{roy2018recalibrating, pereira2019adaptive, rickmann2020recalibrating}).

All models were trained using a mini-batch size of 16. Note that this value is empirically found to be the smallest mini-batch size that ensures that the modules that leverage batch normalization are not at a disadvantage \cite{wu2018group}. The Adam gradient descent algorithm \cite{Kingma14} was used with an initial learning rate of $1\times 10^{-4}$ over 400 epochs. A scheduler adjusted the learning rate by a factor of 0.5 should a plateau of the training procedure be detected with a patience of 10 epochs. We ensured that these fixed experimental parameters allowed all models to converge. All models were evaluated in terms of their Dice-Sørensen score.

\subsection{Results}
\label{sec:experiments:results}

\begin{figure}
    \centering
    \includegraphics[width=1.00\columnwidth]{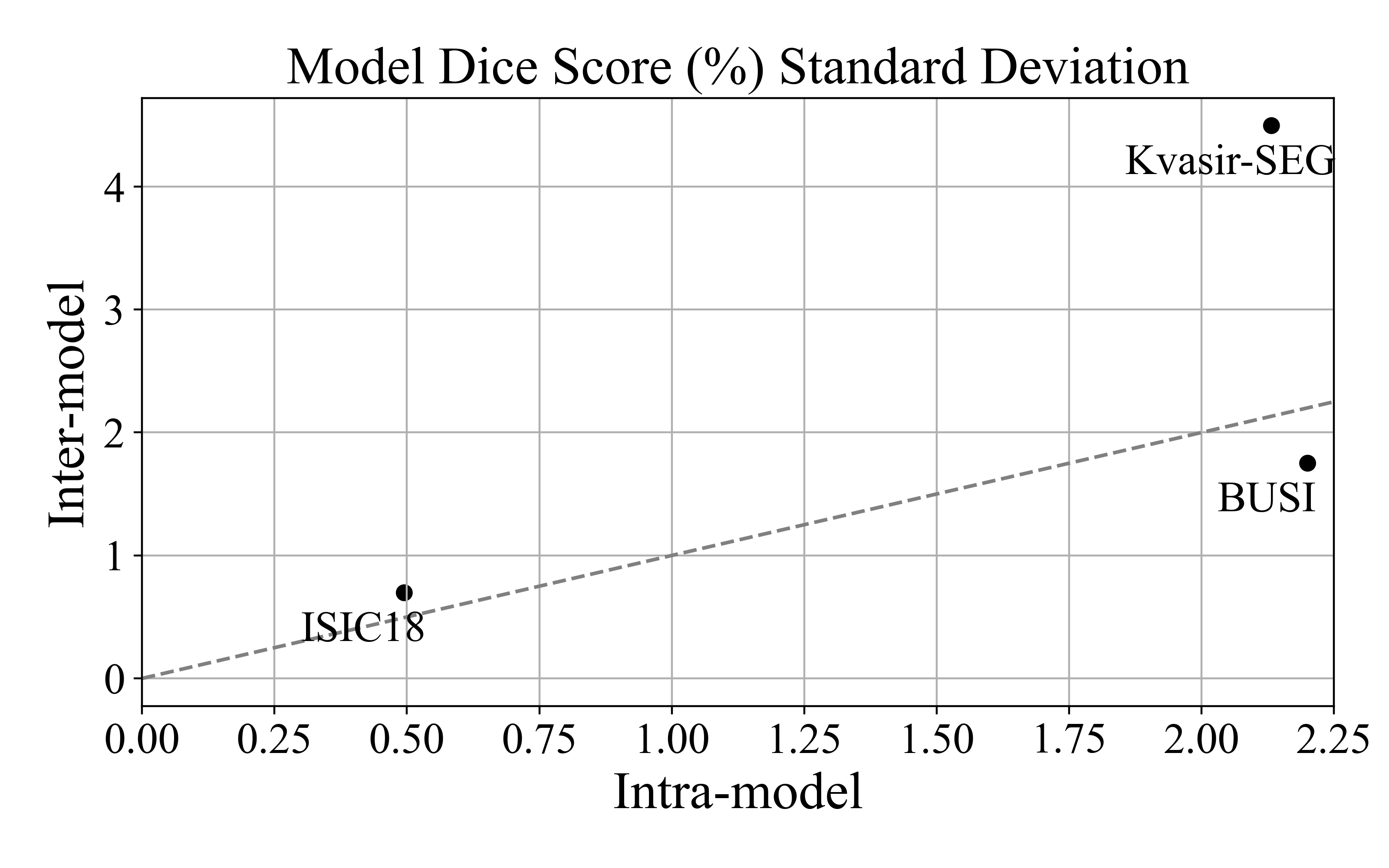}
    \caption{Dice score (\%) inter and intra-model standard deviation per dataset. The dashed gray line marks a ratio of inter/intra model variability of 1.}
    \label{fig:dice_var}
\end{figure}


The 5-fold experiments reveal that at least one of the proposed approaches always outperforms all others for all datasets (see Table \ref{tab:isic_results}). Both cSE and Multifractal recalibration are always beneficial on average when compared to the baseline. However, Multifractal recalibration is the only approach that improves the baseline in a statistically significant manner throughout all datasets. Both of the proposed approaches simultaneously show superior performance to all others in the ISIC18 and BUSI datasets. 

\begin{table}[ht]
\centering
\caption{Mean $\pm$ standard deviation Dice score ($\%$) of the cross-validation experiments. $\cdot^\dag$ and $\cdot^\ddag$ signify that the null-hypothesis of the pairwise t-test with regards to the U-Net baseline is rejected with $p \leq 0.05$ and $p \leq 0.01$, respectively. Best mean results are boldfaced.} 
\begin{tabular}{l c c c}
\hline
Model & ISIC18 & Kvasir-SEG & BUSI \\
\hline
U-Net \cite{ronneberger2015u} & 85.40 $\pm$ 0.25 & 72.22 $\pm$ 1.82 & 62.20 $\pm$ 2.40 \\
+cSE \cite{hu2018squeeze} & 85.94 $\pm$ $0.36^\dag$ & 72.72 $\pm$ $1.52$ & 65.36 $\pm$ 1.36\\
+scSE \cite{roy2018recalibrating} & 85.92 $\pm$ $0.29^\dag$ & 72.94 $\pm$ $1.07$ & 64.82 $\pm$ 1.03 \\
+SRM \cite{lee2019srm} & 84.33 $\pm$ 1.27 & 61.13 $\pm$ 3.42 & 68.09 $\pm$ $3.14^\dag$\\
+FCA \cite{qin2021fcanet} & 86.19 $\pm$ 0.75 & 70.00 $\pm$ 2.51 & 66.27 $\pm$ 2.48 \\

+Mono (ours) & $86.24 \pm 0.27^{\ddag}$ & 71.86 $\pm$ $2.37$ & $\mathbf{69.00 \pm 2.53}^{\ddag}$ \\
+Multi (ours)& $\mathbf{86.26 \pm 0.28}^{\ddag}$ & $\mathbf{74.76 \pm 2.20}^{\dag}$ & 66.94 $\pm$ $2.45^\dag$  \\
\hline
\end{tabular}

\label{tab:isic_results}
\end{table}
It is important to analyse how one should weigh the evidence per dataset, so following \cite{isensee2024nnu} we analyze the ratios of intra an inter-model standard deviation for each dataset (see Figure \ref{fig:dice_var}). Both ISIC18 and Kvasir-SEG are above the threshold where the inter-model variance is of a higher average magnitude than intra-model. BUSI is below this threshold, but it is the only dataset where all of the approaches displayed a dramatic improvement over the baseline, with the highest differences being for SRM and Monofractal recalibration, $+5.89\%$ and $+6.8\%$ Dice, respectively. This dataset is thus less reliable for assessing differences between attention functions, but it is still valid to show that attention functions can improve baseline performance drastically for this task.   

It is noteworthy that our approaches are always superior to FCA, which is the most intricate statistical channel descriptor. SRM also captures second-order information, but it is only competitive in the BUSI dataset, and fails to even match baseline performance in the other cases. 

Monofractal recalibration appears to have greater impact on performance when it works well, compared to cSE and even scSE.  An interesting result is that cSE was superior on average to scSE in the BUSI dataset, and coincidentally, in the same dataset, Monofractal outperformed all other approaches, including Multifractal recalibration. We hypothesize that the speckle noise present in the ultrasound images \cite{yap2008novel} might lead to local singularities that are still propagated by $\Psi_l$, which leads to diminishing returns faster for these spatial approaches. Keep in mind that in spite of this fact, the Multifractal approach still shows sizeable statistical significant improvement over the baseline for this dataset, and that our analysis ranks BUSI lowest  with regards to distinguishing differences between extensions of the baseline U-Net (see Figure \ref{fig:dice_var}). Despite this, Multifractal recalibration still increased baseline Dice score significantly.

Interestingly, Monofractal recalibration alongside the other sophisticated global statistical channel attention functions (SRM and FCA) under-performed when compared not only to cSE and scSE, but also to the baseline. On the other hand, the superior performance of Multifractal recalibration in the same dataset attests to the fact that higher order statistical information is crucial for Kvasir-SEG.

We investigated what kind of information is captured in the singularity map $\mathbf{H}_l$ associated with $\Psi_l$ (see Figure \ref{fig:multifractal-dif}). Qualitatively, the singularities appear to highlight general rugosity patterns latent in the feature maps. Multifractal recalibration effectively clusters this singularity information into macro-structures that become especially visible as $l\rightarrow L$. 


\subsection{Multifractal hyperparameter and aggregation analysis}
\label{sec:experiments:multi-params}
\begin{table}[t]
\caption{Mean $\pm$ standard deviation dice score ($\%$) for values in $Q\in\{2,4,8,16\}$ measured on the ISIC18 dataset. Beast mean results are boldfaced.}
\centering
\begin{tabular}{lllll}
\cline{1-4}
\multicolumn{1}{|l|}{$Q=2$}              & \multicolumn{1}{l|}{$Q=4$}              & \multicolumn{1}{l|}{$Q=8$}              & \multicolumn{1}{l|}{$Q=16$}               &  \\ \cline{1-4}
\multicolumn{1}{|l|}{$86.21 \pm 0.49$} & \multicolumn{1}{l|}{$86.21 \pm 0.32$} & \multicolumn{1}{l|}{$86.15 \pm 0.28$} & \multicolumn{1}{l|}{$\mathbf{86.26 \pm 0.28}$} &  \\ \cline{1-4} 
\end{tabular}

\label{tab:ablate_k}
\end{table}

In order to understand the impact of the number of soft bins in the histogram of Equation \eqref{eq:soft_histogram} we vary the value of the number $Q$ mixtures so that $Q\in\{2, 4, 8, 16\}$. We measure the average and standard deviation of the Dice score in a 5-fold cross-validation setup of the ISIC18 dataset. The results are collected in Table \ref{tab:ablate_k}. Most of the performance is achieved immediately at $Q=2$ histograms. This result becomes less surprising by realizing that this method shares some commonalities with the context encoding proposed in~\cite{zhang2018context}, where $Q$ is always set to the number of classes. However, contrary to~\cite{zhang2018context}, we do not need any additional architectural changes nor loss terms during supervision.
The increase in performance with bigger $Q$ is not significant. However, at $Q=16$ the average performance does increase slightly while simultaneously decreasing the amount of variance. We also found that this value of $Q$ worked well in general for our experiments, as described in Section\ref{sec:experiments:results}.
\begin{table*}[htbp]  
\caption{Mean $\pm$ standard deviation of Multifractal of activation function pooling ablation in the Kvasir-SEG dataset (Dice (\%)). The best mean results are boldfaced.}
\centering
\begin{tabular}{llllll}
\cline{1-5}
\multicolumn{1}{|l|}{$\sum_q \phi(\cdot)$} & \multicolumn{1}{l|}{$\sigma(\sum_q \phi (\cdot))$} & \multicolumn{1}{l|}{$\delta(\sum_q \phi (\cdot))$} & \multicolumn{1}{l|}{$\sigma(\sum_q \delta(\phi (\cdot)))$} & \multicolumn{1}{l|}{$\delta(\sum_q \delta(\phi (\cdot)))$} \\ \cline{1-5}
\multicolumn{1}{|l|}{$73.91 \pm 1.67$} & \multicolumn{1}{l|}{$73.87 \pm 1.38$} & \multicolumn{1}{l|}{$73.66 \pm 1.69$} & \multicolumn{1}{l|}{$\mathbf{74.76 \pm 2.20}$} & \multicolumn{1}{l|}{$69.82 \pm 2.37$} \\ \cline{1-5}
\end{tabular}
\label{tab:aggregation}
\end{table*}

Next, we inquire about the design of the aggregation strategy of $\phi(p_l^{(q)}(\mathbf{H}_l))$ over axis $Q$. These results are collected in Table \ref{tab:aggregation}, which can be broken down into two parts: the choice of (non) linearity inside and outside the pooling aggregation. ReLU is the only candidate inner activation function, since it is already known to be empirically effective \cite{zhang2018context}.

Two surprising facts about our results: the sigmoid in the outer activation function leads to the ultimate performance configuration, and the linear model is the second best in terms of average performance. Concerning the latter, as described in Section \ref{sec:methods:multi}, this aggregation can be seen as a linear mixture of Gaussians, and since $Q=16$ from the previous ablation experiment, this model should already be highly discriminative. However, the effectiveness of the former deserves more careful consideration. Remember that $\Psi_{l}^{\text{Multi}} * \mathcal{F} = (\Psi_{l} * \mathcal{F}) + (\Tilde{\mathbf{H}}_{l}* \mathcal{F})$, where $\mathcal{F}$ can be either a decoder filter at depth $l$ or an encoder filter at depth $(l+1) \leq L$ (we omit $\text{MaxPooling2D}$ here for simplicity). 
The output of this convolution depends linearly on $\Psi_l$ and $\Tilde{\mathbf{H}}_l$, and since $\Psi_l, \Tilde{\mathbf{H}}_l \geq 0$, the sum aggregation bounds the response of this convolution. Choosing the sigmoid makes it so that $\Tilde{\mathbf{H}}_l$ gates the point-wise magnitude of this convolution, similarly to an 
\ac{SE} pathway, but in the spatial dimension. It also prevents the result of this operation from being dominated by $\Tilde{\mathbf{H}}_l$.
 
Finally, we surmise that the effectiveness of the linear model reveals that the proposed aggregation is sensitive to the dying ReLU problem, which may negate the contribution or identification of informative level sets due to initialization of either \eqref{eq:soft_histogram} or $\phi$ \cite{lu2019dying}. Note that also that a general singularity spectrum is well defined for the entirity of the real numbers\cite{hentschel1983infinite}. Architectural tweaks in this module should be further explored in future work.
\subsection{Analysis of Global Channel Attention Functions}
\label{sec:results:empirical_channel}
\begin{figure}[t]
    \centering
    \includegraphics[width=1.00\columnwidth]{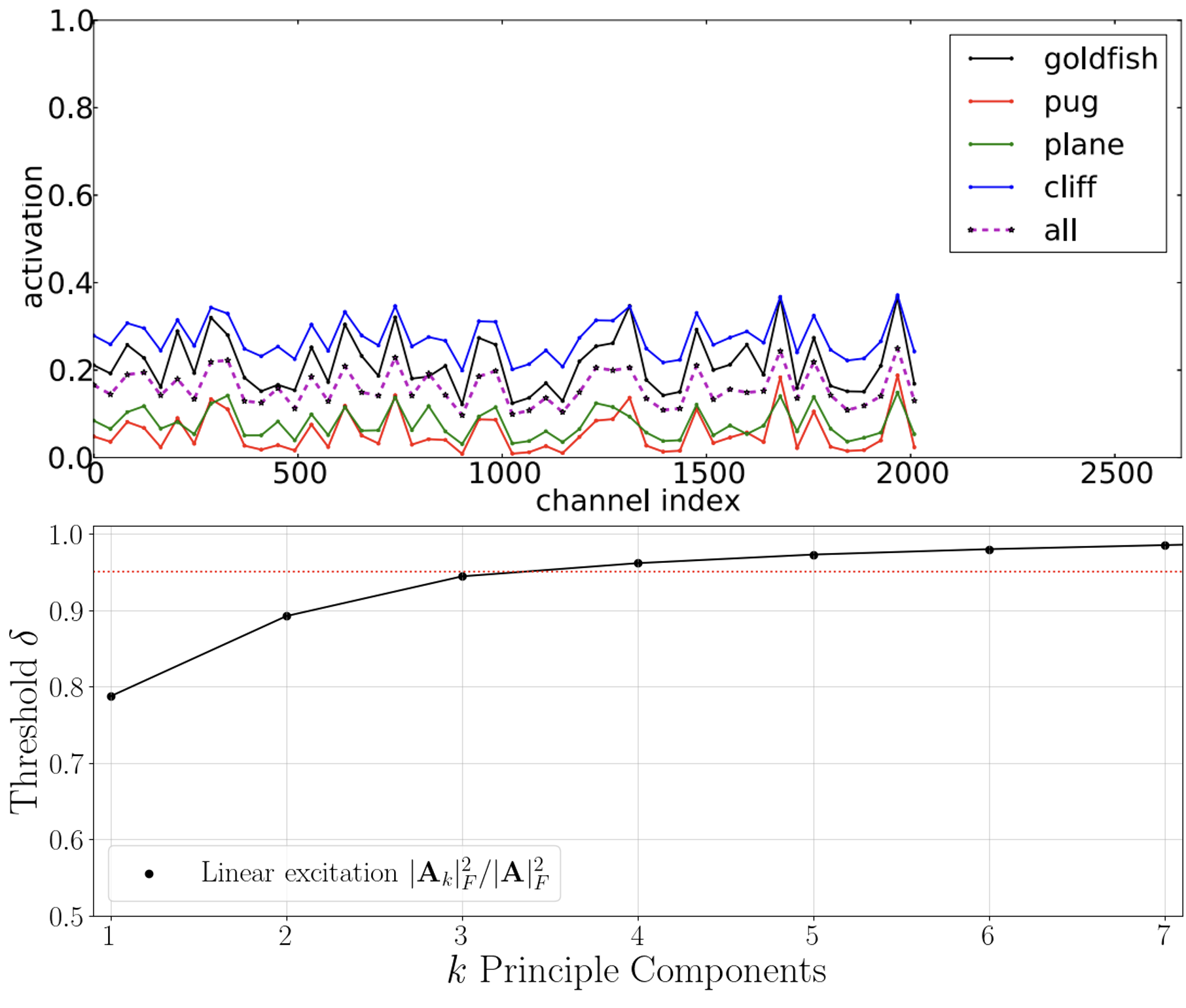}
    \caption{(Top) Excitation analysis of excitation magnitude on a ImageNet-1000 classification task, adapted from \cite{hu2018squeeze}. (Bottom) The number of principal components that capture $\delta=0.95$ of the excitation variance on a binary segmentation task on the ISIC-18 dataset for the first encoder head of a U-Net deprived of its skip-connections.}
    \label{se-cls-seg}
\end{figure}
The authors of the original SE paper \cite{hu2018squeeze} argued, for the task of image classification, that the excitations appeared to be  more class-specific as $l\rightarrow L$, becoming (at least) visually separable before the output layer (see Fig. \ref{se-cls-seg}). Since for segmentation each input location maps to each output, and this relationship becomes impossible to estimate given the lack of spatial resolution of the excitations $g$. 

We use the linear estimation of the covariance of the excitations $g(\psi_l)$. We denote the covariange matrix as $\mathbf{A}\in \mathbb{R}^{C_l\times C_l}$ such that $\mathbf{A} := \frac{1}{n-1}g(\psi_l)^Tg(\psi_l)$, where $n$ is the size of a validation set.
Note that this aligns with recent trends in deep learning theory that state that, under mild assumptions, deep representations tend to display linear behaviour after training \cite{park2023linear, huh2024platonic}.

Since we deal only with binary tasks, we use how much information $\mathbf{A}_k$, the singular value decomposition of $\mathbf{A}$ using only its top $k$ singular values , captures about the variance of the neurons (see Fig. \ref{se-cls-seg}). We define the \textit{linear excitation threshold}, $\delta \in (0, 1]$ as:
\begin{equation}
    \operatorname{argmin}_k \frac{|\textbf{A}_k|^2_F}{|\textbf{A}|^2_F} \geq \delta.
    \label{eq:ept}
\end{equation}
In simple terms, for a fixed $\delta$, if the excitations are linearly related to the target variable, then one expects the minimum $k\rightarrow 2$.

Analysis of this surrogate task with $\delta=0.95$, for each model and dataset pair, yielded three main insights:

\begin{figure}[htbp]
    \centering
    \includegraphics[width=1.00\textwidth]{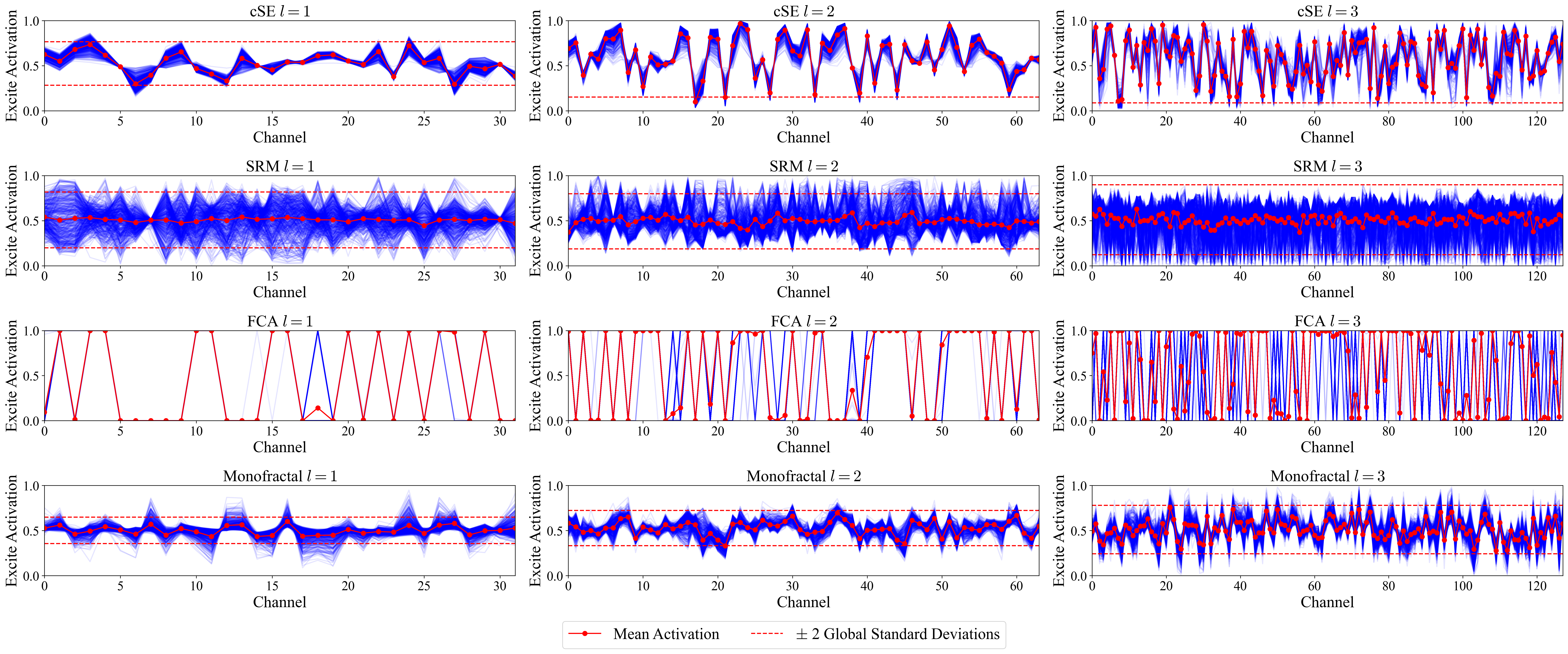}
    \caption{Test set excitation responses of cSE, SRM, FCA, and Monofractal recalibration for $l\in \{1,2,3\}$ in the first fold of the ISIC dataset. Each blue segment marks a response of the layer given an instance. SRM displays very heterogeneous behaviour per instance. On the opposite end of the spectrum, FCA is almost instance-agnostic. cSE and Monofractal recalibration display similar instance variability, but markedly different excitation responses around the average.}
    \label{fig:multifractal-dif-2}
\end{figure}

\begin{figure}[htbp]
    \centering
    \includegraphics[width=\linewidth]{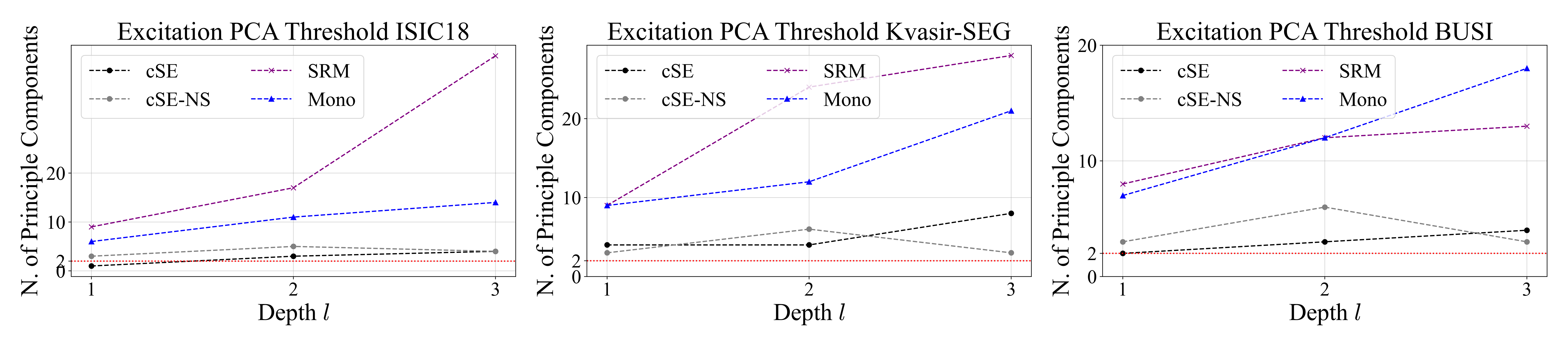}
    \caption{Number of principle components required to explain 95\% of the variance for each model for every encoder depth $l$ for each dataset. cSE-NS stands for cSE in an U-Net without skip-connections.  FCA is not included due to its almost heterogeneous behavior. 
    The U-Shape architecture makes it so that become less class-dependent as $l \rightarrow L$.}
    \label{fig:pca_threshold}
\end{figure}

{\textit{(i) There is no strong evidence of filtering of non-informative channels:}} a common intuition behind channel attention functions is that they somehow learn to select only the most informative channels \cite{roy2018recalibrating, guo2022attention, azad2022medical}. We were not able to measure such an effect in our experiments, at least for the cases when the attention modules actual increase baseline performance. For example, Monofractal recalibration displays the highest performance for the ISIC18 dataset (see Table \ref{tab:isic_results}), although most of the channel are not weighted aggressively by its excitations (see Fig.~\ref{fig:multifractal-dif-2}). 

{\textit{(ii) The covariance of $g$ depends monotonically on $l$ due to its skip-connections:}} we tested measuring the excitation threshold by removing the skip-connections of the U-Net (see Fig.\ref{fig:pca_threshold}, cSE-NS) and found that this would break the monotonic quasi-linear relationship between and $l$ and $k$. A similar behaviour was identified by the original authors of the SE \cite{hu2018squeeze} for classification. Since $C_0 \ll C_l$, this also suggests that the dimension of the linear manifold where the approximation of $\operatorname{Cov}(g)$ resides increases with $l$, for a given $\delta$. This is in line with current intuitions about how more ``complicated features`'' are learned in the later stages of a CNN \cite{zeiler2014visualizing, raghu2017svcca}.

{\textit{(iii) Balanced instance variability appears to correlate with performance:}} extreme instance variability consistently led to subpar performance, as it can be observed for SRM and FCA in Fig.\ref{fig:multifractal-dif-2}. In contrast, both cSE and Monofractal recalibration appear to predictable, although flexible instance variability. 
\begin{table}[t]
\caption{Compute requirements. Inference and train times averaged over 1000 steps using a NVIDIA GeForce RTX 3090 GPU.}
\label{tab:compute}
\centering
\resizebox{\textwidth}{!}{%
\begin{tabular}{|c|c|c|c|c|c|c|c|}
\hline
 & \textbf{U-Net}\cite{ronneberger2015u} & \textbf{+cSE} \cite{hu2018squeeze} & \textbf{+scSE} \cite{roy2018recalibrating} & \textbf{+SRM}~\cite{lee2019srm} & \textbf{+FCA}~\cite{qin2021fcanet} & \textbf{+Mono} & \textbf{+Multi} \\ \hline
\textbf{Total parameters} & 1,946,881 & 1,968,721 & 1,968,948 & 1,947,783 & 3,022,417 & 1,989,105 & 1,967,553 \\ \hline
\textbf{Trainable parameters} & 1,946,881 & 1,968,721 & 1,968,948 & 1,947,335 & 1,968,721 & 1,969,169 & 1,947,521 \\ \hline
\textbf{Non-trainable parameters} & 0 & 0 & 0 & 448 & 1,053,696 & 19,936 & 20,032 \\ \hline
\textbf{Train time (ms/step)} & 60  & 73  & 104  & 77  & 76  & 165  & 450  \\ \hline
\textbf{Inference time (ms/step)} & 40  & 45  & 64  & 50  & 69  & 115  & 220  \\ \hline
\end{tabular}%
}
\end{table}

\section{Discussion and Conclusions}
\label{sec:experiments:discussion}

We introduce two empirical priors based on Fractal Geometry: Monofractal and Multifractal recalibration. We construct them in a principled way and integrate them sucessfully in practice. The proposed channel attention functions significantly enhanced a U-Net's segmentation performance, whilest simultaneously outperforming other statistical informed channel attention functions \cite{roy2018recalibrating, lee2019srm, qin2021fcanet}. Insofar as we are aware, this was the first time that Multifractal analysis was leveraged completely end-to-end for semantic segmentation.

The experiments were conducted on three public medical datasets suggest that although cSE, scSE \cite{roy2018recalibrating} and Multifractal improve the baseline for all datasets, only Multifractal recalibration does so in a statistically significant manner for all cases. 

Our analysis includes other statistical priors that extend traditional SE by including higher order information in some capacity, specifically the SRM \cite{lee2019srm} and FCA \cite{qin2021fcanet}. Surprisingly, they generally underperformed in our experiments. In terms of global channel statistical descriptors, with the exception of the Kvasir-SEG dataset, Monofractal recalibration offers superior performance. 

Our empirical analysis also reveals that global channel attention function effectiveness is potentially linked with instance variability, not to the degree to which these attention models ``filter out'' some channels \cite{hu2018squeeze, woo2018cbam, pereira2019adaptive, guo2022attention}. On the extremes, we observed instances where instance variability was excessive (SRM) or close to 0 (FCA), and in both cases, performance decreased on average. On the other hand, both cSE and Monofractal recalibration offer balanced instance variability but have markedly different average gating responses (refer to Figure \ref{fig:multifractal-dif}). We also demonstrated through experiment that, for the U-Net architecture, the excitation responses seem to get more chaotic with the depth of the encoder, which can be likely attributed to an effect of its skip connections.

We also take a more theoretically grounded approach then previous work in end-to-end (multi)fractal analysis for the context of classification tasks \cite{xu2021encoding, chen2021deep, chen2024mfen}. Notwithstanding, not all limitations were overcomed. For instance, in order to keep computation time reasonable, we had to resort to a simple OLS estimate using a fixed small number of scales. Our recalibration modules require less compute than \cite{xu2021encoding, chen2021deep, chen2024mfen}, yet they still take a considerable amount of extra computational time when compared to the other recalibration strategies studied in this paper (see Table \ref{tab:compute}). However, we achieve a model size that is very close to the baseline U-Net. 

Our inductive prior also relies on the assumption that each filter will define a self-similar measure in order for the OLS step carried out to estimate a quantity that relates to the scaling exponents. We believe some of this problems might be overcomed by further exploring the connections of CNNs with Wavelet Scattering \cite{bruna2013invariant} which potentially will allow to leverage more sophisticated multifractal formalisms \cite{abry2015bridge, jaffard2019multifractal, wendt2022multifractal}.
%
\bibliographystyle{plain}
\bibliography{references}
\end{document}